%% file: main.tex
\documentclass[acmtog,nonacm]{acmart}
\AtBeginDocument{%
  }

\setcopyright{acmcopyright}
\copyrightyear{2018}
\acmYear{2018}
\acmDOI{XXXXXXX.XXXXXXX}

\acmJournal{TOG}
\acmVolume{37}
\acmNumber{4}
\acmArticle{111}
\acmMonth{8}

\input{macros}

\usepackage{graphicx}

\usepackage{booktabs}
\usepackage{ulem}

\usepackage{multirow}
\usepackage{amsmath,bm}
\usepackage{url}
\usepackage[ruled]{algorithm2e} 

\SetAlFnt{\small}
\SetAlCapFnt{\small}
\SetAlCapNameFnt{\small}
\SetAlCapHSkip{0pt}
\usepackage{overpic} 
\usepackage{subcaption}
\usepackage{array}
\usepackage{float}

\usepackage[capitalize]{cleveref}
\crefname{section}{Sec.}{Secs.}
\Crefname{section}{Section}{Sections}
\Crefname{table}{Table}{Tables}
\crefname{table}{Tab.}{Tabs.}

\newcommand{\Fref}[1]{Fig.~\ref{#1}}

\newcommand{\eref}[1]{Equation~(\ref{#1})}

\usepackage{array}
\newcolumntype{L}[1]{>{\raggedright\let\newline\\\arraybackslash\hspace{0pt}}m{#1}}
\newcolumntype{C}[1]{>{\centering\let\newline\\\arraybackslash\hspace{0pt}}m{#1}}
\newcolumntype{R}[1]{>{\raggedleft\let\newline\\\arraybackslash\hspace{0pt}}m{#1}}

\newcommand{\view}{\vv}
\newcommand{\network}{\mathcal{F}}

\long\def\ignorethis#1{}
\definecolor{crimson}{rgb}{0.86, 0.08, 0.24}
\definecolor{green}{rgb}{0, 0.5, 0.25}
\definecolor{purple}{rgb}{0.75, 0, 1}
\definecolor{orange}{rgb}{1, 0.5, 0.25}
\definecolor{yellow}{rgb}{1, 1, 0}
\definecolor{new_blue}{rgb}{0, 0.5, 1}
\definecolor{new_cyan}{rgb}{0.10, 0.62, 0.57}

\begin{document}



\title{Mesh-Guided Neural Implicit Field Editing}


\author{Can Wang}
\email{cwang355-c@my.cityu.edu.hk}
\affiliation{%
  \institution{City University of Hong Kong}
  \country{}
}

\author{Mingming He}
\affiliation{%
  \institution{Netflix Eyeline Studios}
  \country{}
}
\email{hmm.lillian@gmail.com}

\author{Menglei Chai}
\affiliation{%
  \institution{Google AR Perception}
  \country{}
  }
\email{cmlatsim@gmail.com}

\author{Dongdong Chen}
\affiliation{%
 \institution{Microsoft Cloud + AI}
 \country{}
 }
\email{cddlyf@gmail.com}

\author{Jing Liao}
\authornote{Corresponding Author.}
\affiliation{%
  \institution{City University of Hong Kong}
  \country{}
  }
\email{jingliao@cityu.edu.hk}


\begin{abstract}
Neural implicit fields have emerged as a powerful 3D representation for reconstructing and rendering photo-realistic views, yet they possess limited editability. Conversely, explicit 3D representations, such as polygonal meshes, offer ease of editing but may not be as suitable for rendering high-quality novel views. To harness the strengths of both representations, we propose a new approach that employs a mesh as a guiding mechanism in editing the neural radiance field. We first introduce a differentiable method using marching tetrahedra for polygonal mesh extraction from the neural implicit field and then design a differentiable color extractor to assign colors obtained from the volume renderings to this extracted mesh. This differentiable colored mesh allows gradient back-propagation from the explicit mesh to the implicit fields, empowering users to easily manipulate the geometry and color of neural implicit fields. To enhance user control from coarse-grained to fine-grained levels, we introduce an octree-based structure into its optimization. This structure prioritizes the edited regions and the surface part, making our method achieve fine-grained edits to the neural implicit field and accommodate various user modifications, including object additions, component removals, specific area deformations, and adjustments to local and global colors. Through extensive experiments involving diverse scenes and editing operations, we have demonstrated the capabilities and effectiveness of our method. Our project page is: \url{https://cassiepython.github.io/MNeuEdit/}

\end{abstract}


\begin{teaserfigure}
\setlength{\tabcolsep}{0\linewidth}
\includegraphics[width=1.0\textwidth]{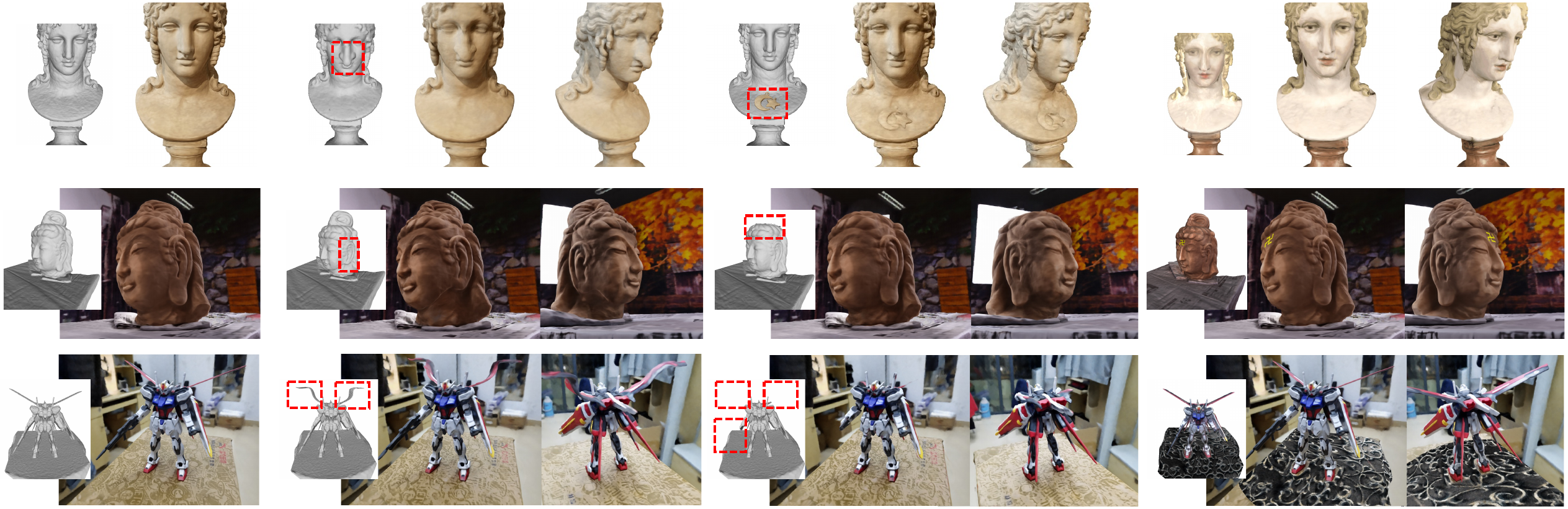}
\begin{small}
\begin{tabular}
{C{0.15\linewidth}C{0.35\linewidth}C{0.15\linewidth}C{0.35\linewidth}}
Source & Geometry deformation & Addition or removal & Texture editing
\end{tabular}
\end{small}
\vspace{-0.6cm}
\caption{Our approach enables the manipulation of both the geometry and color of neural implicit fields through differentiable colored meshes. This includes tasks such as adding or removing geometry, deforming existing geometry, and editing textures. We recommend reviewing the video provided in the supplementary material for more edited results.}
\label{fig:teaser}
\end{teaserfigure}


\maketitle


\section{Introduction}\label{sec:intro}

Today we are witnessing the emergence of Neural Implicit Fields~\cite{mildenhall2020nerf,sitzmann2019scene,liu2020neural} as an emerging content medium revolutionizing the way that is revolutionizing the way humans create and interact with 3D content. Its remarkable ability to model complex 3D scenes and render their photo-realistic novel views has led to its adoption in a wide range of practical applications including VR/AR~\footnote{https://www.lifewire.com/nvidias-instant-nerf-can-turn-your-phots-into-3d-scenes-in-seconds-5224116}, gaming~\footnote{https://neuralradiancefields.io/nerfs-in-unreal-engine-5-alpha-announced-by-luma-ai}, VFX~\footnote{https://www.wrapbook.com/blog/neural-radiance-fields}, among others. With that comes increasing demands from creators to edit the neural implicit fields according to their preferences. However, editing such volumetric representations is challenging due to the implicit encoding of scene appearance within neural features and network weights, which can hardly support intuitive and precise modifications. Recent research has sought to address this by enabling appearance editing of neural implicit fields, guided by various inputs such as an exemplar image~\cite{wang2022clip,bao2023sine,yang2022neumesh,kobayashi2022decomposing,liu2021editing}, a text prompt~\cite{wang2022clip,wang2023nerf,haque2023instruct,bao2023sine,jiang2023avatarcraft,kobayashi2022decomposing}, or a palette~\cite{kuang2023palettenerf,wu2022palettenerf,gong2023recolornerf}. These approaches primarily focus on editing appearance features or color styles of neural implicit fields, or making minor geometry adjustments but do not offer extensive support for non-rigid deformation or topology modification.

\begin{figure}[t]
\setlength{\tabcolsep}{0\linewidth}
\includegraphics[width=\linewidth]{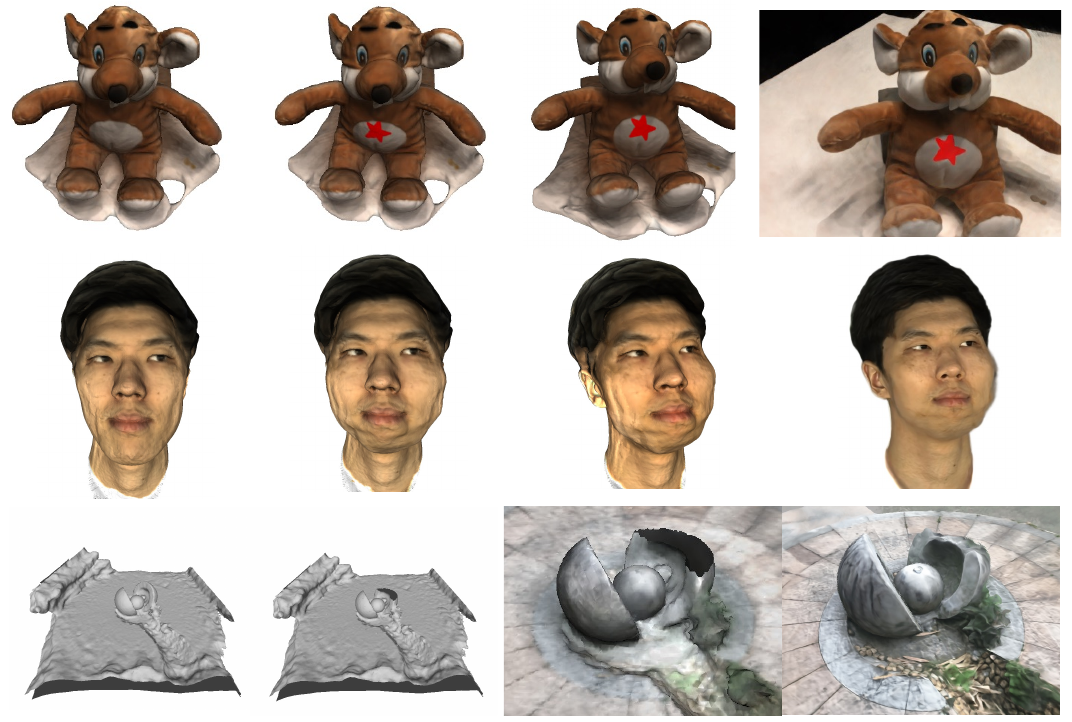}
\begin{small}
\begin{tabular}{p{0.23\linewidth}
<{\centering}p{0.24\linewidth}<{\centering}p{0.24\linewidth}<{\centering} p{0.24\linewidth}<{\centering}}
Source & User Edits & Mesh Rendering & Ours
\end{tabular}
\end{small}
\vspace{-10pt}
\caption{\textbf{The Reasoning Behind Avoiding Direct Rendering of Edited Meshes.} The rendered mesh results are not suitable as final outputs since they lack the ability to handle background, photo-realistic rendering, and intricate details, such as the bear's surroundings, the reality of the human face, and the details of the complex scene.}
\label{fig:reason_no_mesh_render}
\vspace{-10pt}
\end{figure}

A more desirable approach to editing the neural implicit field is to make it as user-friendly as manipulating an explicit mesh in traditional graphics workflows, ensuring compatibility with popular 3D software like Maya, Blender, and more. Unlike a 3D mesh, the neural implicit field lacks an explicit shape for artists to manipulate directly, but it offers the capability of achieving photo-realistic rendering with high-fidelity scene details, a significant difference from the rendering of user-edited 3D meshes. Rendering 3D-edited meshes faces challenges in managing multiple aspects simultaneously, including background integration, achieving photorealism, and preserving intricate details. As shown in Fig.~\ref{fig:reason_no_mesh_render}, this struggle becomes evident in scenarios like depicting the bear's surroundings, capturing human facial reality, and rendering complex scene intricacies. Hence, we are motivated to leverage the advantages of both 3D representations by employing a mesh to guide the editing process within the neural radiance field.

Bridging the gap between the 3D mesh and the neural implicit field to propagate arbitrary changes from explicit to implicit representations is a non-trivial task. Pioneering efforts~\cite{yuan2022nerf,yang2022neumesh} establish correspondence between mesh vertices and neural implicit fields, enabling changes in vertices to affect the implicit field. In one method category, exemplified by~\cite{yuan2022nerf}, the focus is on geometry deformation, achieved by extracting a guidance mesh from the density field using marching cubes and computing a translation field based on vertex alterations. Applying this translation field effectively deforms the geometry, but it struggles with modeling topology changes or supporting color editing.  The other category, as represented by~\cite{yang2022neumesh}, implements mesh-guided neural implicit field editing in a feed-forward manner, encoding mesh vertex positions and colors as inputs for the neural field. This forward-based approach implements geometry deformation through adjustments to vertex positions and facilitates color editing by optimizing texture codes based on user edits. However, it also faces challenges with topology changes and multi-view color consistency, especially for complex textures, due to the single-view guidance. It is worth noting that all these methods share a common limitation: they rely on the non-differentiable mesh representation to update the neural implicit field, limiting their support for topology changes and precise color editing. Consequently, we propose to solve this challenge via a backward-based approach centered on a differentiable colored mesh, allowing for updates to the neural implicit field to accommodate changes in geometry, color, and topology.

We present the first differentiable mesh-guided method for editing neural implicit fields, featuring the extraction of a colored mesh from the implicit field in a differentiable way. This approach enables users to interact with multiple facets of the mesh, encompassing geometry, topology, and color attributes, and back-propagates these updates to the implicit field. We aim to capture user manipulations, extending beyond basic color and mesh edits to offer fine-grained control, thereby ensuring that the rendered results faithfully align with the users' creative intent. To accomplish this, we tackle two challenging problems: how to extract a differentiable colored mesh and how to support different levels of user control. Firstly, drawing inspiration from creative differentiable mesh extraction techniques used in 3D reconstruction~\cite{liao2018deep,gao2020learning,munkberg2022extracting,shen2021deep}, we employ a differentiable marching tetrahedra method~\cite{munkberg2022extracting} for high-quality mesh extraction from the neural implicit field and further extend this representation to extract mesh color differentiably. The proposed differentiable color extractor approximates vertex colors from the radiance integral near the surface during volume rendering, making the process differentiable, much like the geometry extraction. As shown in Fig.~\ref{fig:teaser}, with this differentiable colored mesh, we gain the capability to update the entire neural implicit field and achieve various editing possibilities, such as adding objects, removing sections, or deforming specific areas. Additionally, we can change the color attributes of the implicit fields and consistently render novel views that align with the user's desired changes.

Furthermore, to enhance the differentiable colored mesh and make it support user edits from coarse-grained to fine-grained, we introduce an octree-based structure into the optimization process. When dealing with a target colored mesh, we utilize an octree to create an irregular 3D discrete scalar field, where the scalar density dynamically adjusts in response to user edits, resulting in denser scalar fields for edited areas. Subsequently, we extract a polygonal mesh iso-surface from this density scalar field and apply a process similar to our differentiable color extraction to assign vertex colors to this mesh. By minimizing the Chamfer distance~\cite{fan2017point} between the source extracted mesh and the target edited mesh and optimizing the density MLP layers, we enable modifications to the geometry or topology of a neural implicit model, even for detailed structures. Additionally, this octree-based structure enables us to do comprehensive control over the color of the neural implicit field, like mapping delicate colors to the object surface as shown in Fig.~\ref{fig:teaser}.

In summary, our key idea involves the joint differentiable extraction of both geometry and color within one framework, then applied to the editing of neural implicit fields to align with the user input. In detail:

\begin{itemize}
\item We introduce a unified framework that integrates geometry, topology, and color editing for neural implicit fields, offering users the same ease of manipulation of neural implicit fields as explicit colored meshes. This framework enables extensive flexibility to accommodate various user edits, such as object additions, component removals, specific area deformations, and precise color editing.
\item We present a differentiable method for extracting colored meshes from neural implicit fields, which enables gradient back-propagation from the mesh to the neural implicit fields. Building on this advancement, we propose an octree-based optimization technique to facilitate fine-grained editing of neural implicit fields in both geometry and color.
\item We integrate various training techniques into the editing process, including geometric regularization, vertex augmentation, camera augmentation, and a coarse-to-fine training strategy. These methods improve the octree-based optimization process, resulting in visually compelling outcomes.
\end{itemize}


\section{Related Work}

\begin{figure*}[htbp]
\centering 
\includegraphics[width=\textwidth]{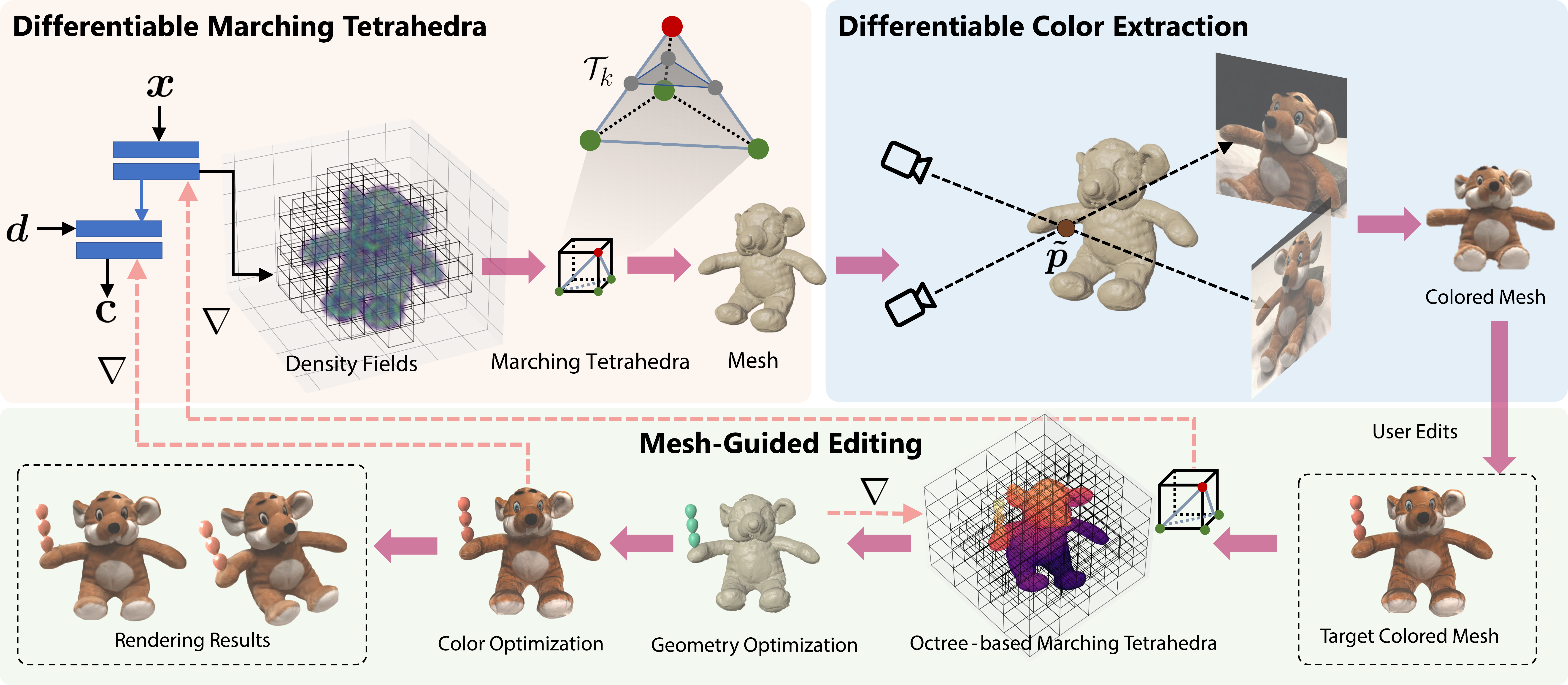}
\vspace{-15pt}
\caption{\textbf{Framework.} To allow editing neural radiance fields with mesh guidance, our method first introduces a differentiable marching tetrahedra for mesh extraction and a differentiable color extraction to produce colors for this mesh. Then given a target mesh, 
our differentiable octree-based colored mesh allows users to edit the neural implicit fields in a mesh-based workflow via a two-step optimization method, such as first optimizing the density fields to add a candy to the mesh and followed by optimizing the radiance fields to paint colors.}
\label{fig:framework}
\vspace{-5pt}
\end{figure*}

\subsection{Neural Implicit Fields Editing} Neural implicit fields are a potent representation for modeling complex 3D scenes and enabling free-view photo-realistic rendering, spurring extensive research in high-quality 3D reconstruction~\cite{jain2021putting,yu2021pixelnerf,zhang2022fdnerf, chen2022mobilenerf,muller2022instant,chen2022tensorf}, and 3D asset generation~\cite{niemeyer2021giraffe,schwarz2020graf, park2021nerfies,park2021hypernerf,poole2022dreamfusion,jain2022zero}.
However, these existing neural implicit models offer limited user control over geometry and color within implicit fields.

Recently, the conditional neural implicit field has been proposed with a latent space conditioned on additional input such as user scribbles~\cite{liu2021editing}, 3DMM parameters~\cite{hong2022headnerf,athar2022rignerf,zheng2022avatar}, or a text prompt~\cite{wang2022clip}. While this technology enables geometry editing by changing the latent code, the dependency on latent space and the requirement for large-scale category-specific training data make it only suitable for very specialized scenarios.
EditNeRF~\cite{liu2021editing} designs a conditional implicit architecture by injecting a shape code and an appearance code that allows editing using user scribes at the image level. The editing is performed by finetuning the network with this user scribe's supervision. CLIP-NeRF~\cite{wang2022clip} first designed a disentangled conditional implicit fields architecture and then introduced CLIP to guide the editing with a text prompt.
However, these methods still lack support for editing implicit fields from a 3D perspective, a recognized, more user-friendly, and intuitive interaction mode for traditional computer graphics users. 
In contrast, NeuTex~\cite{xiang2021neutex} allows users to edit appearance by editing 2D UV maps.
It explicitly learns a mapping from the spatial positions to 2D UV colors through a circle loss. This loss is defined via AtlasNet~\cite{groueix2018papier} that maps 2D UV coordinates into 3D locations, which struggles to reconstruct complex shapes. Additionally, editing desired positions might be challenging due to distorted UV mapping~\cite{yang2022neumesh}. 

An alternative, more intuitive, and flexible approach compared to the aforementioned methods would involve editing the implicit fields using an explicit mesh as a guide. NeRF-Editing~\cite{yuan2022nerf} explicitly defines a deformation field by calculating the vertex translations between the original mesh extracted from a pre-trained model and the target mesh edited by users. Users can render the edited results with this deformation field by deforming the original mesh template. However, this method only supports geometry deformation and does not allow topology changes and color manipulations. NeuMesh~\cite{yang2022neumesh} establishes a forward process that encodes implicit representations into a mesh-based format, featuring disentangled geometry and color codes assigned to mesh vertices. This setup enables mesh-guided geometry editing and specific color alterations such as swapping, filling, and painting. However, such a mesh-based representation for volumetric neural rendering requires a mesh-based scaffold as input, introducing an additional initialization step for mesh extraction. Its geometry editing capabilities are confined to deformations similar to NeRF-Editing which cannot support topology changes. Moreover, NeuMesh restricts color painting to a 2D perspective and requires users to paint on a single view to guide the color code modifications. 
NeUVF~\cite{ma2022neural} expands upon NeuTex's concept~\cite{xiang2021neutex} by improving texture editing and mapping the dynamic NeRF colors onto a 2D texture. For geometry editing, NeUVF utilizes semantically rich facial landmarks to generate a 3D deformation field. As a result, users have the ability to modify the head's shape by manipulating these keypoints. However, akin to NeuTex, NeUVF encounters difficulties linked to distorted UV mapping while editing the desired texture. Furthermore, NeUVF shares with NeRF-Editing~\cite{yuan2022nerf} the inability to facilitate user edits involving topology changes. Unlike existing approaches, our method directly modifies the mesh extracted from implicit fields. We introduce a differentiable colored mesh for editing neural implicit fields, enabling us to facilitate topology changes and color editing within the extracted 3D mesh. This offers users an intuitive, interactive control via a 3D mesh, ensuring a 'what you see is what you get' experience.

\subsection{Differentiable Mesh Extraction}
Marching cubes~\cite{lorensen1987marching} generates a polygonal surface representation of an iso-surface derived from a discrete scalar field. This technique is commonly employed as a post-processing step to extract a polygonal mesh from the density fields produced of a neural implicit model. However, Marching cubes lacks differentiability, preventing gradient propagation to voxels.

To address this limitation,~\citet{liao2018deep} devise a neural network to predict voxel occupancy probability and edge vertex displacement. This approach enables defining a mesh distribution differentiably using the predicted occupancy probability and vertex locations. The network is integrated as a final layer within a 3D convolutional network to learn the prediction of 3D meshes.
Similar to~\citet{liao2018deep}'s work, DefTet~\cite{gao2020learning} 
also utilizes a neural network to predict a vertex displacement and the occupancy for each tetrahedron. 
At its essence, the approach optimizes concurrently to predict both vertex displacement and tetrahedral occupancy for 3D reconstruction, operating within a pre-defined tetrahedral mesh structure.
However, the computational cost escalates cubically with the grid resolution of this occupancy representation, restricting its capability to represent high-resolution 3D shapes effectively. 

DMTet~\cite{shen2021deep} improves DefTet by representing the geometry using a sign distance field (SDF) instead of the occupancy. This SDF is also defined on a deformable tetrahedral grid as DefTet did. DMTet applies selective tetrahedra subdivision around the predicted surface to enhance representation capabilities while maintaining manageable computational complexity. NVDiffrec~\cite{munkberg2022extracting} also leverages the differentiable marching tetrahedrons from DMTet for topology optimization. The difference is that NVDiffrec introduces a differentiable renderer to support PBR materials and environment map lighting jointly learned from 2D supervision. 
Inspired by DMTet and NVDiffrec, we use the differentiable marching tetrahedrons to represent mesh extracted from neural implicit fields for the future editing. But our approach sets itself apart from both methods by extending the original differentiable marching tetrahedrons through a novel octree-based method, which allows for fine-grained geometry editing. 
Furthermore, DMTet and NVDiffrec lack support for color extraction in neural implicit fields, limiting user capabilities for color space modifications. To address this, we introduce an innovative differentiable color extraction technique that enables color assignment to vertices. This advancement permits users to modify mesh colors and effectively propagate these alterations back to the neural implicit fields.


\section{Method}\label{sec_mtd}

\newcommand{\netdensity}{\mathcal{F}_{\sigma}}
\newcommand{\netcolor}{\mathcal{F}_c}

We proposed a mesh-guided neural implicit field editing method based on differentiable mesh representations. Before diving into these differentiable representations, we first review the general formulation of neural implicit fields (\S~\ref{sub_NeRF}). Then, we introduce the specifics of the differentiable representation for colored meshes. This includes differentiable marching tetrahedra for polygonal meshes extracted from neural implicit fields (\S~\ref{sub_DMT}), alongside differentiable vertex color extraction used to colorize the extracted meshes (\S~\ref{sub_DCE}). This process supports gradient back-propagation from the mesh to the neural implicit fields. Additionally, we designed an octree-based optimization method for finely detailed neural implicit fields in geometry and color editing (\S~\ref{sub_GCE}), followed by the implementation of various editing techniques in geometry and color. The schematic illustration of the framework is shown in Fig.~\ref{fig:framework}

\subsection{Neural Implicit Fields}\label{sub_NeRF}
Neural implicit fields~\cite{mildenhall2020nerf,oechsle2021unisurf,wang2021neus,yariv2021volume} represent a 3D scene by defining a continuous field as an implicit function. This function $\network$, parameterized by two MLP blocks $\netdensity$ and $\netcolor$, maps a spatial position $\vx(x,y,z)$ together with a view direction $\view(\phi,\theta)$ to a density $\sigma$ and a view-dependent radiance $\vc$ respectively. The neural implicit model renders novel views via a volume rendering:
\begin{equation}
    \mathrm{C}(\vr) = \int_{t_n}^{t_f} w(t)\vc(\vr(t), \vd)dt
    \label{eq:volume_integration}
\end{equation}
where the ray $\vr$ from the camera origin $\vo$ follows $\vr(t) = \vo + t\vd$ and $w(t)$ is the weight decided by $\sigma(\vr(t))$.
Without loss of generality, we adopt NeuS~\cite{wang2021neus} as our base neural implicit model due to its merit in geometry representation.

\begin{figure}[t]
\includegraphics[width=0.47\textwidth]{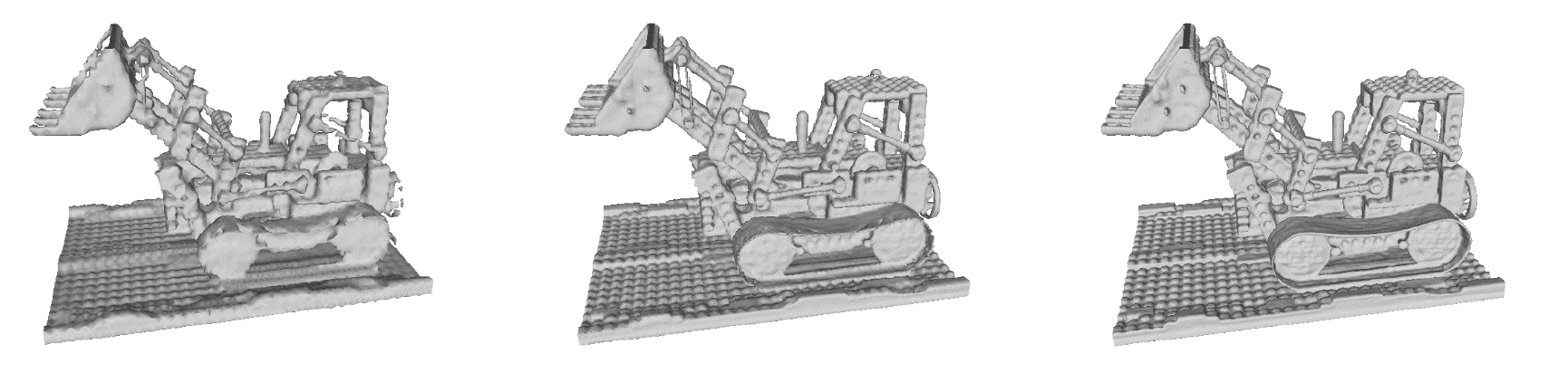}
\begin{small}
\begin{tabular}{p{0.28\linewidth}<{\centering}p{0.28\linewidth}<{\centering}p{0.28\linewidth}<{\centering} }
N=64 & N=128 & N=256
\end{tabular}
\end{small}
\vspace{-15pt}
\caption{\textbf{Mesh Extraction Comparison.} The differentiable marching tetrahedra produces more accurate geometry with a higher grid resolution.
}
\label{fig:abla_mc}
\end{figure}

\subsection{Differentiable Marching Tetrahedra}\label{sub_DMT}

Since editing the neural implicit field, represented by an implicit function, can be challenging, we propose using mesh guidance to enhance the editing process. Our first goal is to extract a mesh from the neural implicit field.
Inspired by~\cite{shen2021deep,munkberg2022extracting} which uses differentiable marching tetrahedra~\cite{shen2021deep,munkberg2022extracting} for geometry reconstruction, we adapt this representation to extract geometry from neural implicit fields in a differentiable manner.
Given the pre-trained implicit function $\network=\netdensity\circ 
 \netcolor$, we first determine which locations are occupied by the object. To achieve this, we create a 3D regular grid volume 
$\mathcal{V} \subseteq \mathbb{R}^{3\times N \times N \times N}$
in the form of a cuboid. To make this grid volume cover the whole object, we restrict its values between the maximum and minimum boundary values $b_\mathrm{max}$ and $b_\mathrm{min}$ of the bounding box. 
Intuitively, a larger
value of $N$ induces a more concise mesh. 
$\netdensity$ takes $\mathcal{V}$ as input and outputs a scalar density grid $\mathcal{D} \subseteq \mathbb{R}^{N\times N\times N}$. 
We denote the position of a point in $\mathcal{D}$ as $\vv_i=(x_i,y_i,z_i)$, where $i\in\left \{ 1,\cdots,N^3 \right \}$ and $x_i,y_i,z_i\in[0,N-1]$. 
Then we split
each cube in $\mathcal{D}$ into five regular tetrahedra and denote all tetrahedra as $\mathbf{T}$. Each tetrahedron $\mathcal{T}_k\in \mathbf{T}$ is represented with its four points $\left\{\vv_{a_k},\vv_{b_k},\vv_{c_k},\vv_{d_k}\right\}$ and four corresponding density values $\left\{\sigma_{a_k},\sigma_{b_k},\sigma_{c_k},\sigma_{d_k}\right\}$. 
Afterward, similar to marching cubes in extracting mesh from neural implicit models, we define a threshold $s$ to indicate the sign of $\vv_i$:
\begin{equation}
    \sign(\vv_i)=\left\{\begin{matrix}
+1  , \, \sigma_i > s,\\ 
-1 , \, \sigma_i \leq s.
\end{matrix}\right.
    \label{eq:sign}
\end{equation}
Then the surface typology inside $\mathcal{T}_k$ can be identified.
In an edge of $\mathcal{T}_k$, a vertex is placed in case of a sign change of density values of two adjacent positions in an edge.
The vertex location of the iso-surface is defined by the
zero crossings of the linear interpolation:
\begin{equation}
    {\vv}'_{ab}=\frac{\vv_a\cdot \sigma_b-\vv_b\cdot \sigma_a}{\sigma_b-\sigma_a}.
    \label{eq:vertex_position}
\end{equation}
We process all cubes in $\mathcal{D}$ and obtain the vertex positions $\mathbf{P}$ and faces $\mathbf{F}$ to produce the triangle mesh $\mathcal{M}=(\mathbf{P},\mathbf{F})$. As \eref{eq:vertex_position} is only evaluated when $\sign(\vv_a)\neq \sign(\vv_b)$, the iso-surface extraction process is differentiable. 

We show mesh extraction results at different resolutions from the differentiable marching tetrahedra in \Fref{fig:abla_mc}. The higher grid resolution achieves more accurate geometry. At the resolution of $128$ and $256$, the differentiable marching tetrahedra produces an accurate mesh topology that preserves geometric features such as sharp edges and smooth curves.

\subsection{Differentiable Color Extraction}\label{sub_DCE}

The differentiable mesh extraction enables geometry editing; however, it does not support color editing since the vertices do not have assigned colors.
Based the extracted mesh $\mathcal{M}=(\mathbf{P},\mathbf{F})$, we design a differentiable color extraction method to assign colors for $\vp\in\mathbf{P}$. Here vertices are normalized into the bounding box as $\frac{\vp}{N-1}(b_\mathrm{max}-b_\mathrm{min})+b_\mathrm{min}$.

\begin{figure}[t]
\centering
\includegraphics[width=0.4\textwidth]{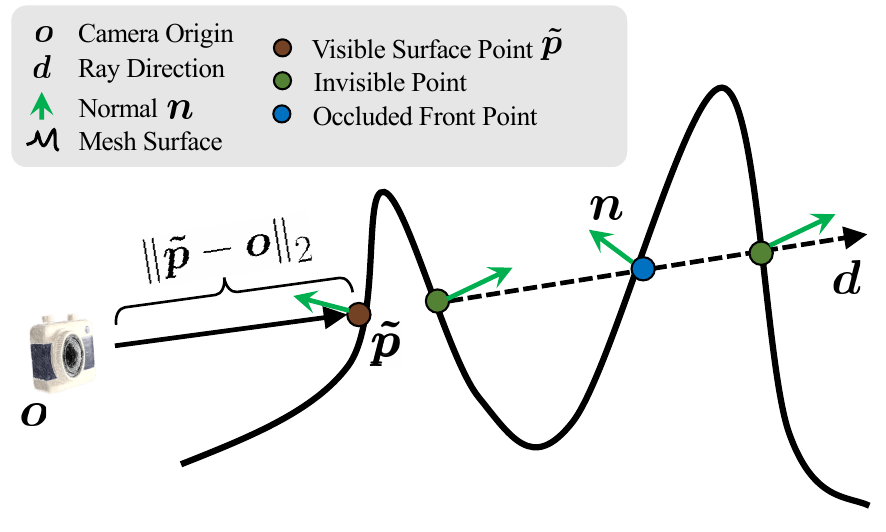}
\vspace{-5pt}
\caption{\textbf{Surface Point Visibility Test in Differentiable Color Extraction.} We filter invisible and occluded front points using normal and depth test. }
\label{fig:method_vc}
\vspace{-10pt}
\end{figure}

Given a camera origin $\vo$ outside the bounding box, we define the ray directions along $\vo$ to $\vp$ as $\vd=\frac{\vp-\vo}{\left\|\vp-\vo\right\|}$. 
We decide the visibility of $\vp$ with regard to $\vo$ by checking whether $\vn\cdot\vd<\mathbf{0}$, where $\vn$ denotes the vertex normal. 
Then we have visible surface points $\tilde{\vp}\in\tilde{\mathbf{P}}$.
However, this simple solution can not filter out occluded front-facing vertices, as shown in \Fref{fig:method_vc}.
Intuitive solutions are ray marching~\cite{perlin1989hypertexture} or Möller–Trumbore intersection algorithm~\cite{moller2005fast}. However, both of them are slow when processing large numbers of rays and vertices. Instead, we adopt the depth test to remove occluded vertices. Specifically, we first derive the depth value $\tilde{d}$ on the ray with $\tilde{\vd}=\frac{\tilde{\vp}-\vo}{\left\|\tilde{\vp}-\vo\right\|}$ from $\netdensity$ and calculate the distance from $\vo$ to $\tilde{\vp}$ as $||\tilde{\vp}-\vo||_2$. Ideally, if $\tilde{\vp}$ is a visible surface point, $\tilde{\vd}$ should be close to $||\tilde{\vp}-\vo||_2$. Thus we remove vertex $\tilde{\vp}$ when $|\tilde{\vd}-||\tilde{\vp}-\vo||_2|>\epsilon$, with the threshold $\epsilon$ set to a small value ($0.2$ in all our experiments). We finally derive occlusion-aware visible surface points $\hat{\mathbf{P}}$ after removing all the occluded vertices from $\tilde{\mathbf{P}}$.

Recall the volume rendering in \eref{eq:volume_integration}, to derive vertex colors for $\hat{\mathbf{P}}$, we have:
\begin{equation}
    \mathrm{C}(\hat{\vr}) = \int_{t_n}^{t_f} w(t)\vc(\hat{\vr}(t),\hat{\vd})dt.
    \label{eq:volume_integration_color}
\end{equation}
As $\hat{\vp}\in\hat{\mathbf{P}}$ under multiple camera views may have common vertices, we average $\mathrm{C}(\hat{\vr})$ under all camera origins to obtain final vertex colors for $\hat{\vp}$.

\begin{figure}[t]
\centering
\includegraphics[width=0.5\textwidth]{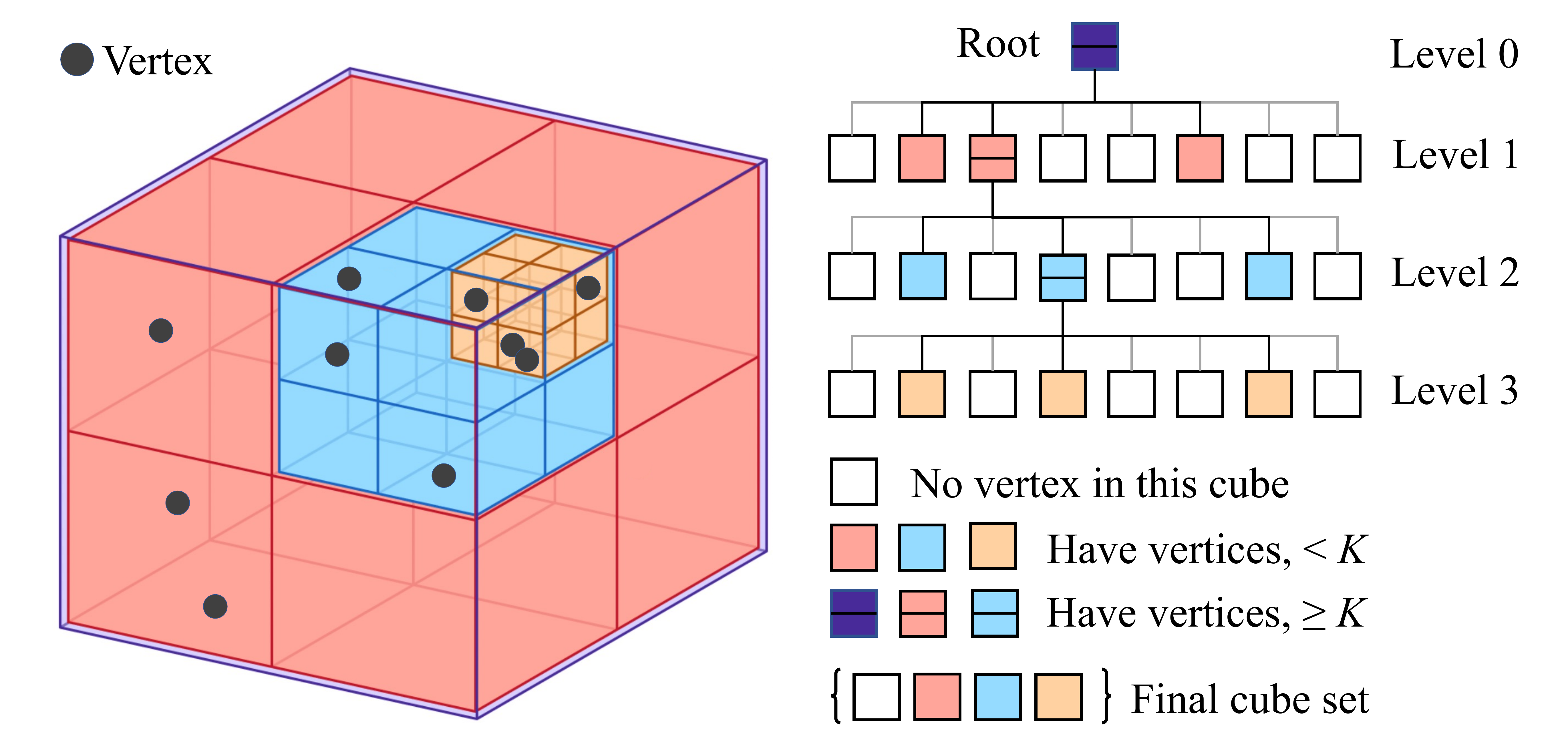}
\vspace{-5pt}
\caption{\textbf{Illustration of our octree via a toy example.} Given vertices from a target mesh, we partition a cube by checking whether its vertex number $\geq K$. We iteratively construct this octree until no cube meets the partition condition or reaching the maximum depth level. The final cube set will be used to extract mesh.}
\label{fig:method_octree}
\vspace{-10pt}
\end{figure}

\subsection{Mesh-guided Editing}\label{sub_GCE}
The process of utilizing the aforementioned differentiable marching tetrahedra to extract a mesh from the density grid $\mathcal{D} \subseteq \mathbb{R}^{N\times N\times N}$ is characterized by being both time-consuming and demanding in terms of memory usage. This is primarily due to the significantly large value of $N^3$, which leads to heightened computational expenses and often triggers out-of-memory (OOM) issues when propagating gradients across each grid within the density field. Consequently, fine-grained editing of the density fields becomes unfeasible, necessitating an exceedingly dense grid for such modifications. Similar constraints are encountered in the context of finely adjusting colors, as achieving adequate colored vertices for color optimization also demands a dense grid. Notably, a substantial portion of the density field regions do not contribute to the geometry, emphasizing the need to focus more on the object surface. Consequently, we propose an optimization approach based on octrees for mesh and color optimization after user editing to reduce computational cost and enable fine-grained editing.

\noindent{\textbf{\textit{Octree-based Optimization}}}.
We first freeze the neural implicit fields and extract the source mesh $\mathcal{M}_\mathrm{s}=(\mathbf{P},\mathbf{F})$ as above and construct the octree based on $\mathbf{P}$. In \Fref{fig:method_octree}, we initialize a root cube at level zero $\mathcal{S}_0$ covering $\mathbf{P}$ and then set a threshold $K$ and check whether $\left | \mathcal{S}_0 \right |\geq K$ (where $\left | \mathcal{S}_0 \right |$ represents the vertex number). If true, this parent node will be subdivided into eight sub nodes at level one $\mathcal{S}_1^i,i\in \left [1,8  \right ]$. We will then check whether $\left | \mathcal{S}_1^i \right | \geq K$ to decide the continuous subdivision. Iteratively, we obtain the final octree structure and the final cube set, represented by a 3D irregular grid $\mathcal{V}^{octree}$. $\netdensity$ takes $\mathcal{V}^{octree}$ as input and outputs an irregular density grid $\mathcal{D}^{octree}$. We finally extract the new mesh $\mathcal{M}^{octree}=(\mathbf{P}^{octree},\mathbf{F}^{octree})$ from $\mathcal{D}^{octree}$ using \eref{eq:sign} and \eref{eq:vertex_position}. We move forward to assign vertex colors for $\mathbf{P}^{octree}$ using \eref{eq:volume_integration_color}. We set this colored mesh as a proxy in our editing process.

\begin{figure}[t]
\includegraphics[width=0.34\textwidth]{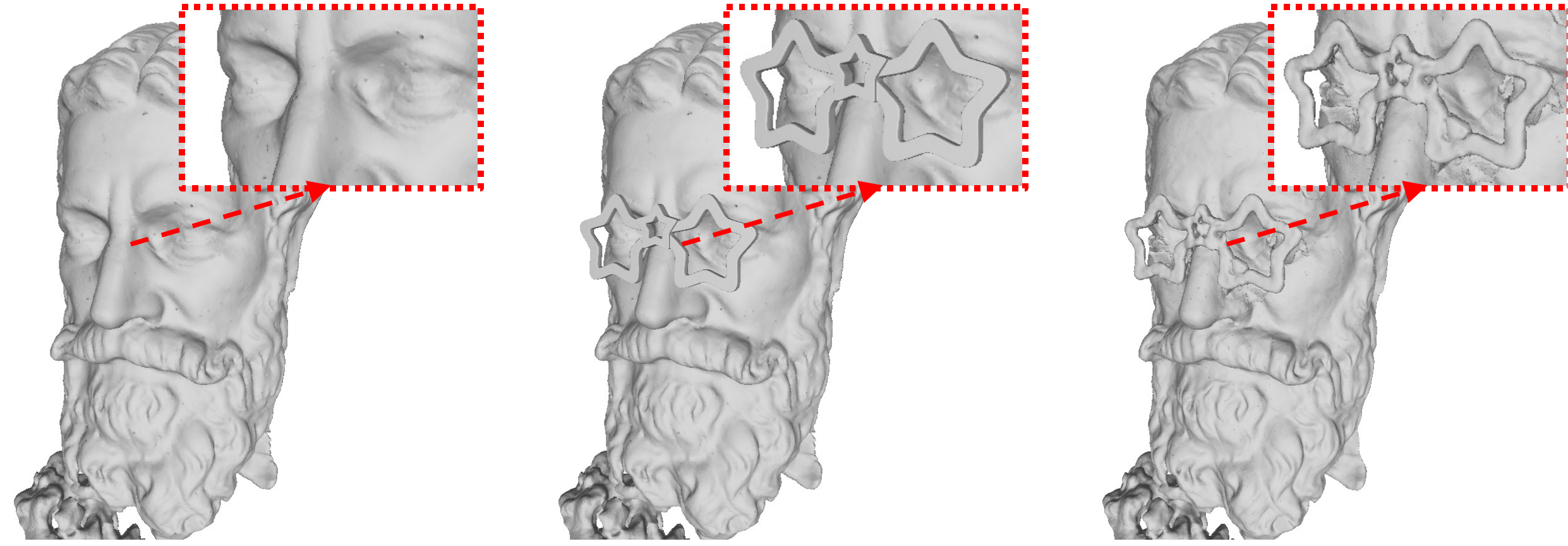}
\begin{small}
\begin{tabular}{p{0.21\linewidth}<{\centering}p{0.20\linewidth}<{\centering}p{0.24\linewidth}<{\centering} }
Source & User Edit & w/o Reg.
\end{tabular}
\end{small}
\includegraphics[width=0.34\textwidth]{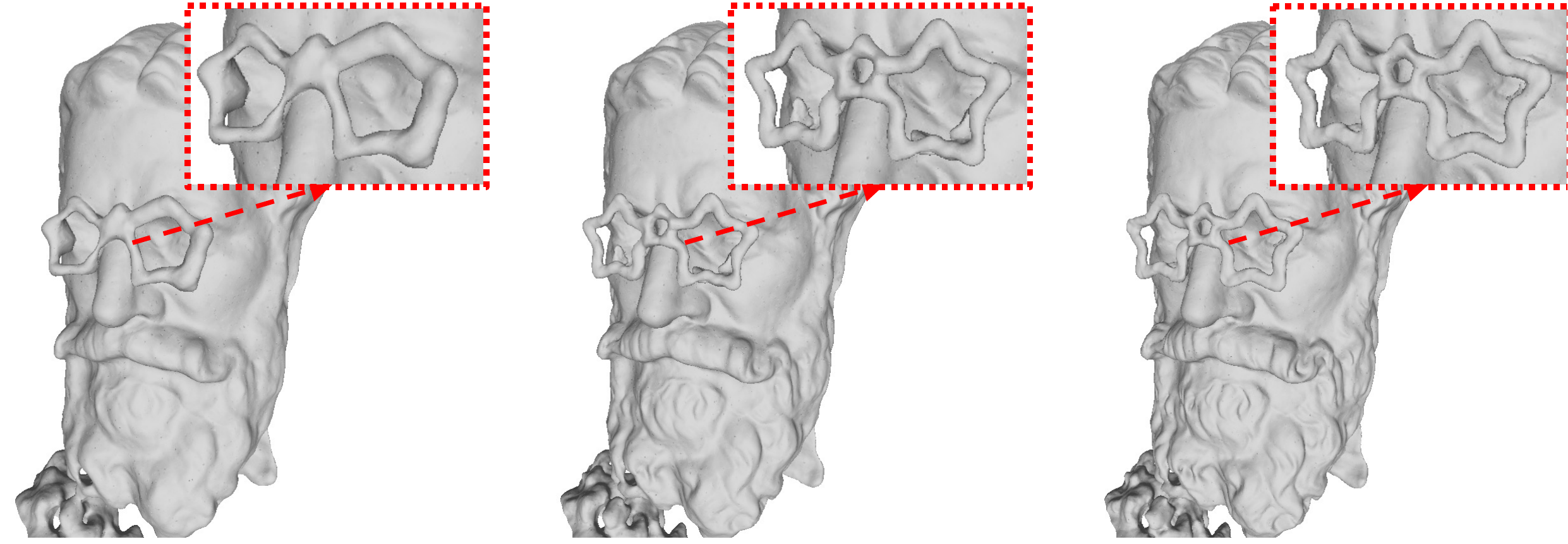}
\begin{small}
\begin{tabular}{p{0.21\linewidth}<{\centering}p{0.20\linewidth}<{\centering}p{0.24\linewidth}<{\centering} }
0.01 & 0.001 & 0.0001
\end{tabular}
\end{small}
\vspace{-15pt}
\caption{\textbf{Regularization Term Evaluation.} We present our findings by utilizing the regularization term with various weights, as well as comparing the results obtained without employing this term.
Among the different weights tested, we have chosen a weight of 0.0001 for the regularization term, which demonstrates significant improvements in terms of enhancing the smoothness of the optimized mesh and reducing the presence of noisy points. This weight choice consistently favors the generation of a more refined and optimized mesh, compared to the method without employing this term.}
\label{fig:regular}
\end{figure}

\begin{figure*}[ht]
\setlength{\tabcolsep}{0.4\linewidth}
\includegraphics[width=1.0\textwidth]{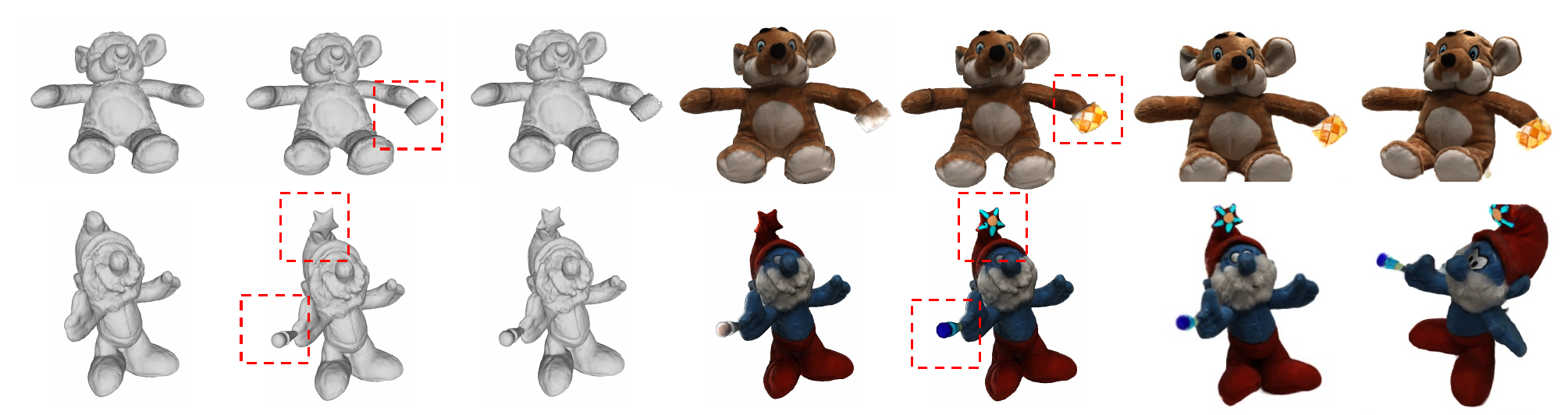}
\begin{small}
\setlength{\tabcolsep}{0\linewidth}
\begin{tabular}{p{0.143\linewidth}<{\centering}p{0.143\linewidth}<{\centering}p{0.143\linewidth}<{\centering}p{0.143\linewidth}<{\centering}p{0.145\linewidth}<{\centering} p{0.283\linewidth}<{\centering} }
Source Mesh & Geometry Editing & Optimized Mesh & Colored Mesh & Color Editing & Rendered Results
\end{tabular}
\end{small}
\vspace{-15pt}
\caption{\textbf{Two Step Optimization.} Our framework enables incremental user editing workflow with different operations: editing the geometry first, then painting the color, and finally rendering the results. }
\label{fig:continue}
\vspace{-10pt}
\end{figure*}

\begin{figure}[t]
\setlength{\tabcolsep}{0\linewidth}
\includegraphics[width=0.94\linewidth]{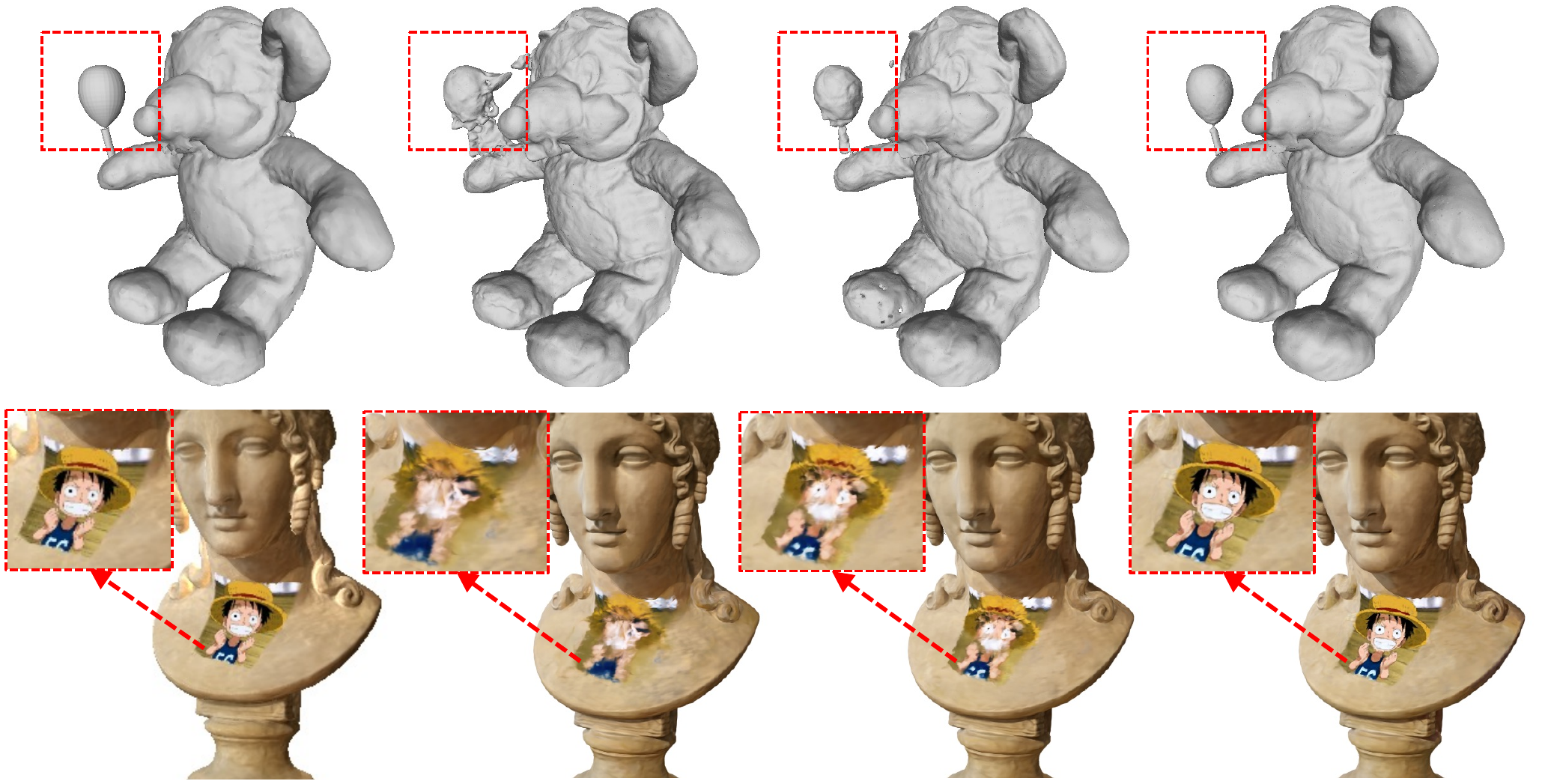}
\begin{small}
\begin{tabular}{p{0.23\linewidth}<{\centering}p{0.23\linewidth}
<{\centering}p{0.23\linewidth}
<{\centering} p{0.23\linewidth}<{\centering}}
User edits & w/o, $N=64$ & w/o, $N=128$ & w/ Octree
\end{tabular}
\end{small}
\vspace{-10pt}
\caption{\textbf{Ablation on Octree.} Our octree-based marching tetrahedra technique empowers more refined geometry and texture editing compared to the non-octree method, across different grid resolutions ($N=64$ and $N=128$). We demonstrate object addition on the bear and texture editing on the sculpture. }
\label{fig:abla_octree}
\vspace{-10pt}
\end{figure}

\begin{figure}[t]
\setlength{\tabcolsep}{0\linewidth}
\includegraphics[width=\linewidth]{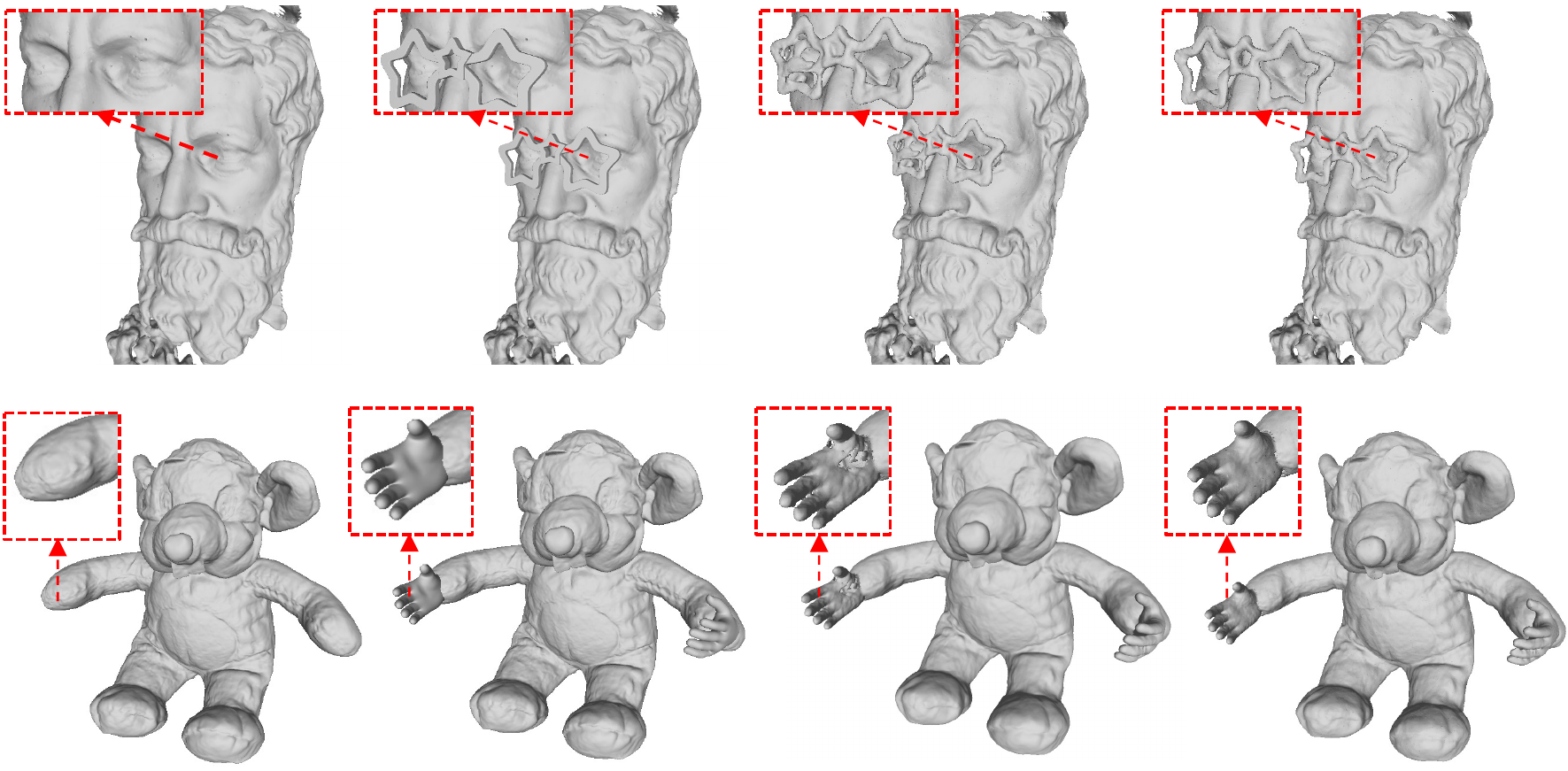}
\begin{small}
\begin{tabular}{p{0.25\linewidth}<{\centering}p{0.25\linewidth}<{\centering}p{0.25\linewidth}<{\centering} p{0.25\linewidth}<{\centering}}
Source & User edits & w/o C2f. & w/ C2f.
\end{tabular}
\end{small}
\vspace{-10pt}
\caption{\textbf{Ablation on Coarse-to-fine Geometry Optimization.} Our coarse-to-fine optimization facilitates smooth convergence of the Chamfer loss and leads to significantly improved geometry optimization results.}
\label{fig:abla_c2f}
\vspace{-10pt}
\end{figure}

\begin{figure}[t]
\setlength{\tabcolsep}{0\linewidth}
\includegraphics[width=\linewidth]{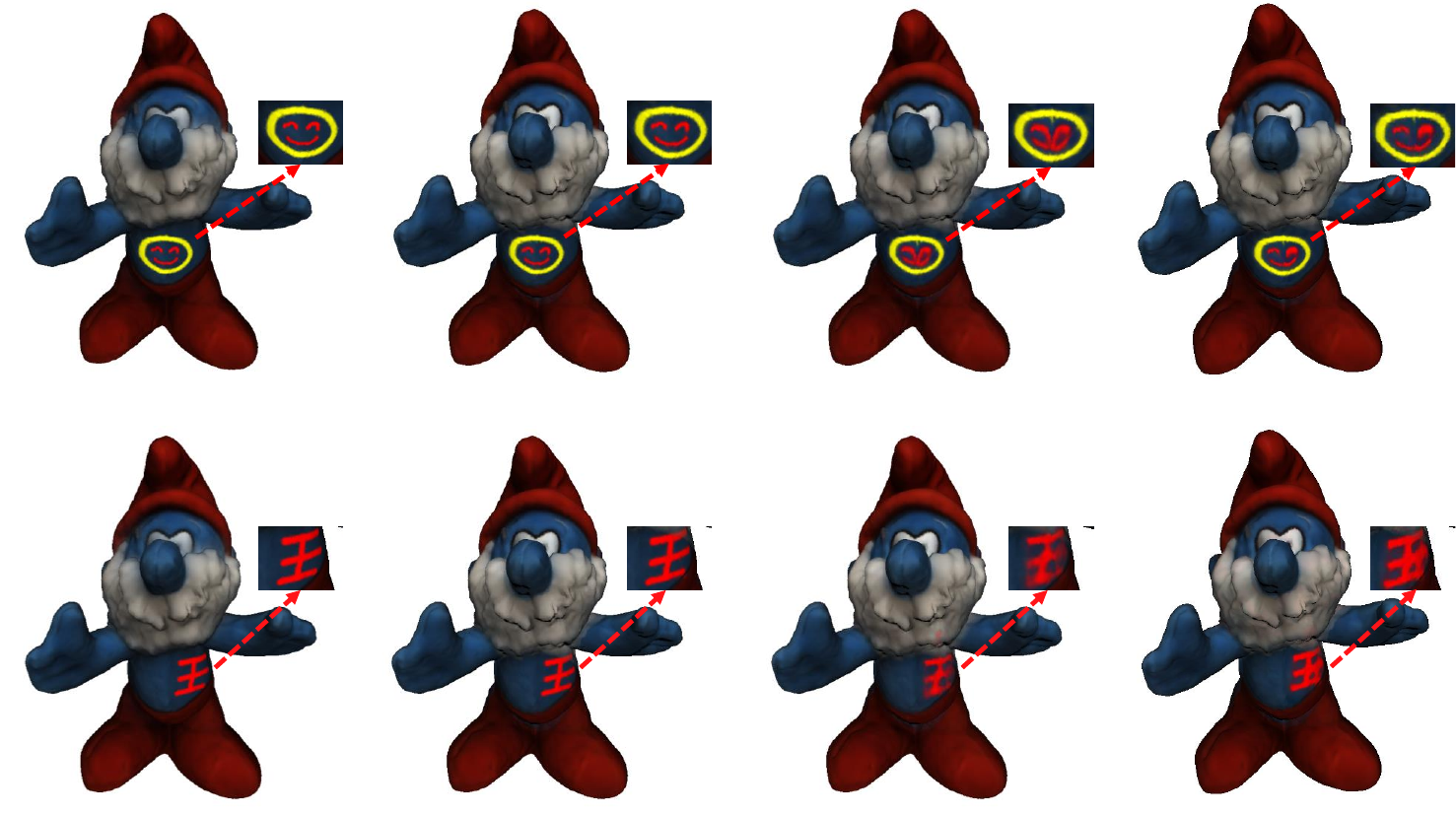}
\begin{small}
\begin{tabular}{p{0.25\linewidth}<{\centering}p{0.25\linewidth}<{\centering}p{0.25\linewidth}<{\centering} p{0.25\linewidth}<{\centering}}
User edits & w/ all Aug. & w/o V. Aug. & w/o C. Aug.
\end{tabular}
\end{small}
\vspace{-10pt}
\caption{\textbf{Ablation on Vertex and Camera Augmentations.} Our vertex and camera augmentations help obtain fine-grained editing results on the optimized mesh while removing these augmentations leads to coarser results.}
\label{fig:abla_va}
\vspace{-10pt}
\end{figure}

\begin{figure}[t]
\includegraphics[width=0.49\textwidth]{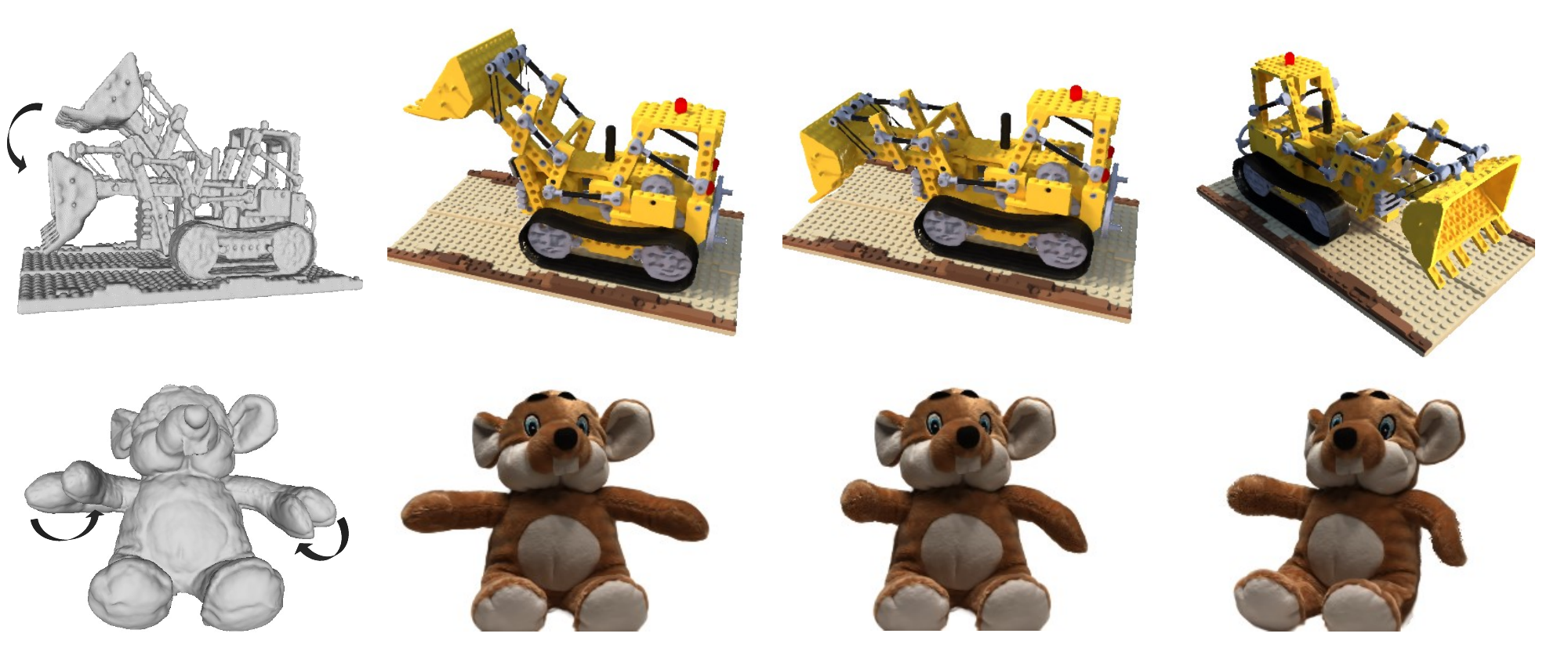}
\begin{small}
\begin{tabular}{p{0.2\linewidth}<{\centering}p{0.2\linewidth}<{\centering}p{0.5\linewidth}<{\centering} }
User Edits & Source & Rendered Results
\end{tabular}
\end{small}
\vspace{-10pt}
\caption{\textbf{Mesh-Guided Object Motion.} We can perform mesh-guided object motions by defining a deformation field calculated from the source mesh to the target one while the topology is fixed. }
\label{fig:result_motion}
\vspace{-10pt}
\end{figure}

\begin{figure*}[ht]
\centering
\setlength{\tabcolsep}{0.4\linewidth}
\begin{small}
\setlength{\tabcolsep}{0\linewidth}
\begin{tabular}{p{0.10\linewidth}<{\centering}p{0.11\linewidth}<{\centering}p{0.26\linewidth}<{\centering}p{0.30\linewidth}<{\centering} p{0.25\linewidth}<{\centering} }
 & Object Motion &   & Color Editing & 
\end{tabular}
\end{small}
\includegraphics[width=0.90\textwidth]{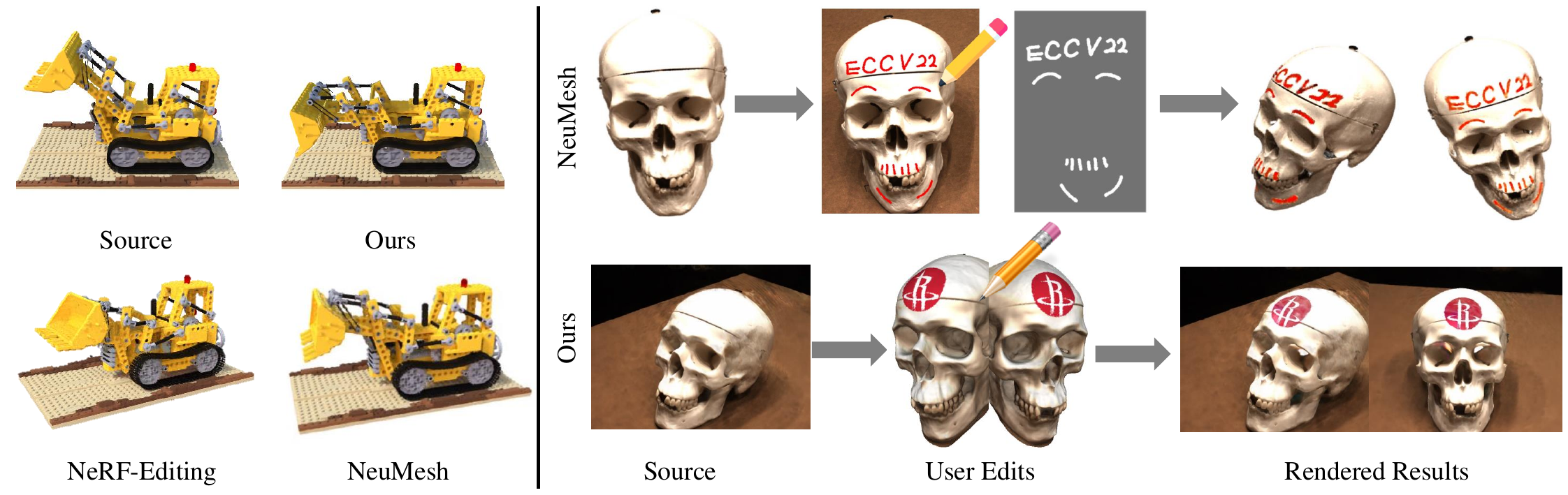}
\vspace{-5pt}
\caption{\textbf{Comparisons with NeRF-Editing~\cite{yuan2022nerf} and NeuMesh~\cite{yang2022neumesh}.} For geometry editing (\textit{Left}), NeRF-Editing and NeuMesh only allow object deformation without topology changes, while our method supports both object motions and topology editings. For color editing (\textit{Right}), we compare our differentiable color editing workflow with NeuMesh. NeuMesh requires users to paint in a single view to provide a mask for color code optimization under 2D supervision. In contrast, our framework allows users to directly edit the 3D color of a mesh, which is more intuitive and accurate. }
\label{fig:cmp}
\vspace{-10pt}
\end{figure*}

\begin{figure*}[ht]
\setlength{\tabcolsep}{0.4\linewidth}
\includegraphics[width=0.97\textwidth]{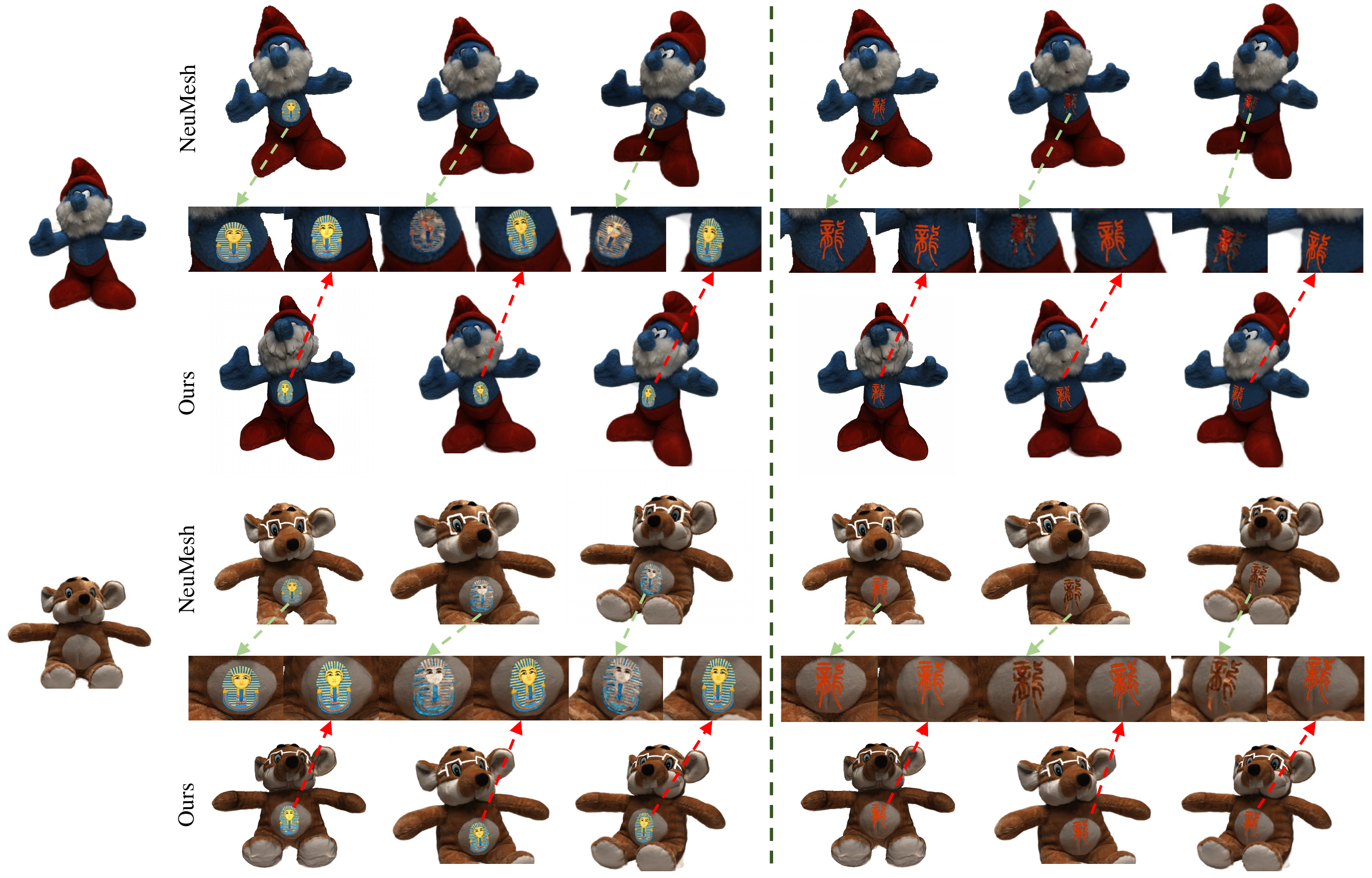}
\begin{small}
\setlength{\tabcolsep}{0\linewidth}
\begin{tabular}{p{0.133\linewidth}<{\centering}p{0.147\linewidth}<{\centering}p{0.284\linewidth}<{\centering}p{0.169\linewidth}<{\centering}p{0.270\linewidth}<{\centering}}
Source & User Edits & Rendered Results & User Edits & Rendered Results
\end{tabular}
\end{small}
\vspace{-15pt}
\caption{\textbf{Visual Comparisons with NeuMesh~\cite{yang2022neumesh}.} Our method outperforms NeuMesh by producing more fine-grained and consistent editing results.}
\label{fig:visualcmp}
\vspace{-10pt}
\end{figure*}

\begin{figure}[t]
\includegraphics[width=0.47\textwidth]{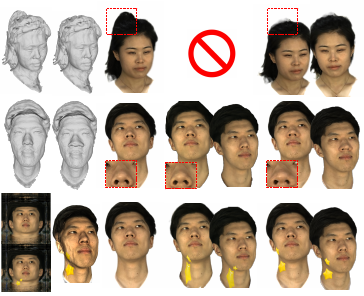}
\begin{small}
\begin{tabular}{p{0.4\linewidth}<{\centering}p{0.24\linewidth}<{\centering}p{0.24\linewidth}<{\centering} }
Source \& User Edits & NeUVF & Ours
\end{tabular}
\end{small}
\caption{\textbf{Visual Comparison with NeUVF~\cite{ma2022neural}.} NeUVF does not accommodate user editing involving topology changes. When it comes to geometry deformation, NeUVF doesn't yield precise results. Additionally, during color editing, distortions may arise because of the distorted UV space.
}
\label{fig:cmp_neuvf}
\end{figure}

\begin{figure*}[t]
\setlength{\tabcolsep}{0.4\linewidth}
\includegraphics[width=0.84\textwidth]{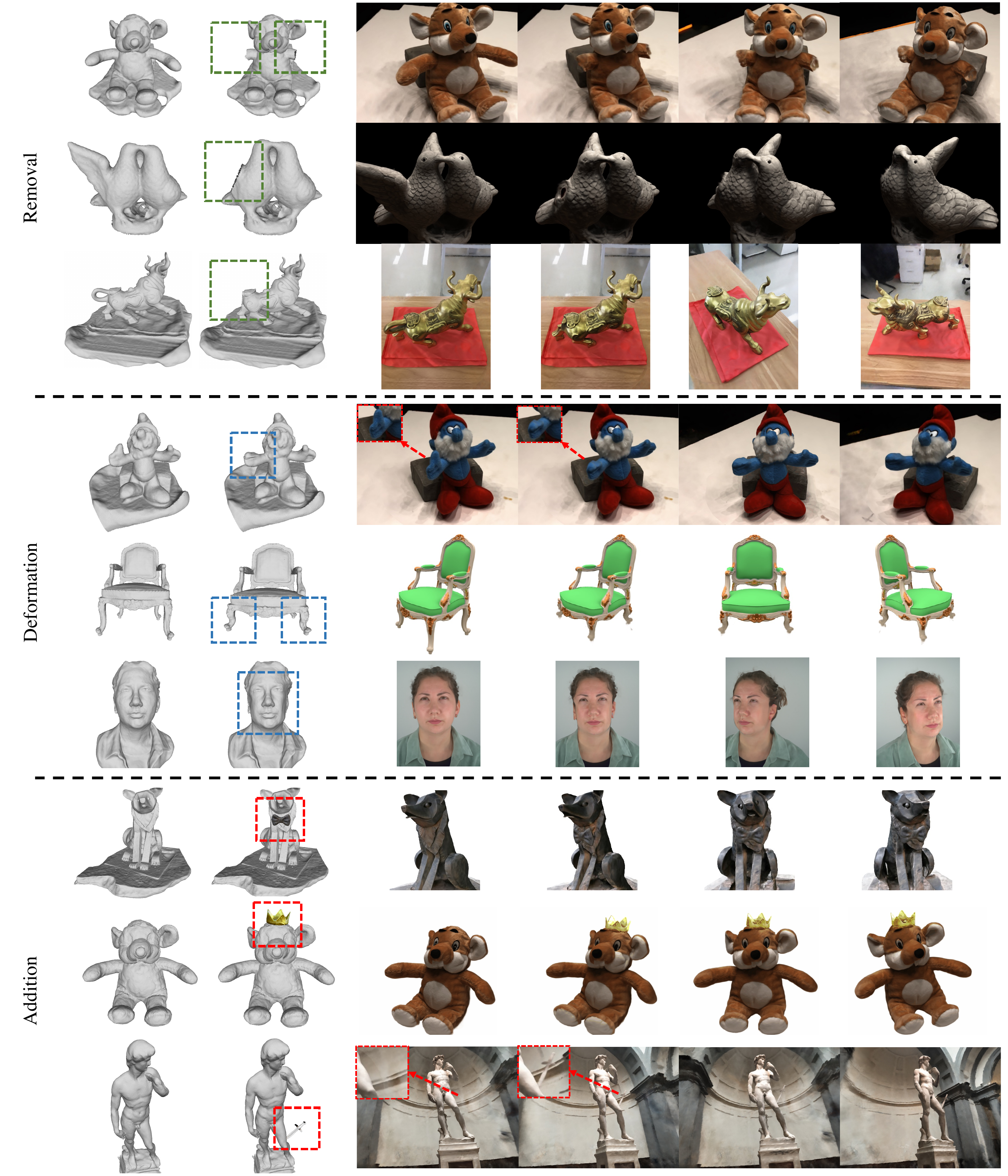}
\begin{small}
\setlength{\tabcolsep}{0\linewidth}
\begin{tabular}{p{0.14\linewidth}<{\centering}p{0.10\linewidth}<{\centering}p{0.14\linewidth}<{\centering}p{0.12\linewidth}<{\centering} p{0.46\linewidth}<{\centering} }
 & Source Mesh  & User Edits & Source & Rendered Results
\end{tabular}
\end{small}
\vspace{-5pt}
\caption{\textbf{Geometry Topology Editing.} Our method supports various geometry edits, including geometry removal~(in green), geometry deformation~(in blue), and adding a new object~(in red).}
\label{fig:ge}
\vspace{-10pt}
\end{figure*}

Our octree-based training exhibits subtle differences depending on the user-editing actions. In the scenarios where a new object is added, we directly construct the octree from the target user-edited mesh, denoted as $\mathcal{M}_\mathrm{t}$. Conversely, when the action involves removing parts from $\mathcal{M}_\mathrm{s}$, we construct the octree from $\mathcal{M}_\mathrm{s}$ rather than $\mathcal{M}_\mathrm{t}$, placing emphasis on the regions being removed and the surface part.
For deformations applied to $\mathcal{M}\mathrm{s}$, the octree is constructed from the intersection of $\mathcal{M}_\mathrm{s}$ and $\mathcal{M}_\mathrm{t}$, represented as $\mathcal{M}_\mathrm{s} \cup \mathcal{M}_\mathrm{t}$. Our approach facilitates dual focus to simultaneously address removed regions and newly introduced ones. While adjusting color attributes, we initiate subdivision within the specified color-editing areas, resulting in the generation of the user-defined target colored mesh, designated as $\mathcal{M}_\mathrm{t}$. In practice, we have the capability to directly build the octree from $\mathcal{M}\mathrm{s} \cup \mathcal{M}\mathrm{t}$, as it encapsulates all the diverse user edits mentioned earlier.

When working with a target colored mesh, our mesh-guided editing follows a two-step optimization process. Initially, we optimize the density function $\netdensity$ to facilitate geometry editing. Subsequently, we optimize the color function $\netcolor$ to achieve color editing.

\noindent{\textbf{\textit{Geometry Editing}}}. Since our geometry extraction is tightly aligned to the density field, any manipulation of the mesh geometry will directly affect the density field and the rendered result. Therefore, users can manipulate a neural implicit model by editing its mesh, such as removing undesired vertices and faces, adding new objects, or deforming certain parts.   

Given the source mesh $\mathcal{M}_\mathrm{s}$ extracted from a pre-trained model $\network=\netdensity\circ \netcolor$ and the target mesh $\mathcal{M}_\mathrm{t}$ edited by users. We first construct $\mathcal{M}^{octree}$ from $\mathcal{M}_\mathrm{s} \cup \mathcal{M}_\mathrm{t}$ and then fine-tune $\netdensity$ with a Chamfer distance:
\begin{equation}
    \mathrm{D}_\mathrm{cd}(\mathbf{S}_1,\mathbf{S}_2|\mathcal{M}_\mathrm{s},\mathcal{M}^{octree})=\sum _{\vx\in\mathbf{S}_1}\min_{\vy\in\mathbf{S}_2}\left\|\vx-\vy\right\|_2^2+\sum_{\vy\in\mathbf{S}_2}\min_{\vx\in\mathbf{S}_1}\left\|\vx-\vy\right\|_2^2,
    \label{eq:chamfer}
\end{equation}
where $\mathbf{S}_1$ and $\mathbf{S}_2$ are vertex sets sampled from $\mathcal{M}_\mathrm{s}$ and $\mathcal{M}^{octree}$, respectively. Since our mesh extraction is differentiable, the gradient can propagate back to $\netdensity$. We finally obtain the optimized $\hat{\netdensity}$, which allows for rendering the edited geometry with the new neural implicit model $\hat{\network}=\hat{\netdensity}\circ \netcolor$. 

To avoid producing noisy points, we include a Eikonal term~\cite{gropp2020implicit} to regularize the gradient with respect to the optimized points in $\mathbf{S}_1$:
\begin{equation}
    \mathcal{L}_\mathrm{reg}=\sum_{x\in \mathbf{S}_1} (\left \| \triangledown \netdensity(x) \right \|-1)^2.
    \label{eq:color}
\end{equation}
We evaluate this regularization term in \Fref{fig:regular}.

We encountered challenges with Chamfer loss convergence at first, especially when dealing with an extensive number of vertices, leading to a significantly large solution space. To mitigate this, we design a strategy involving coarse-to-fine geometry optimization. We extract three source meshes $\mathcal{M}_\mathrm{s}^L$ from $\network$ at three octree depth levels $L=7,8,9$. Finally, we gradually optimize $\mathrm{D}_\mathrm{cd}(\mathbf{S}_1,\mathbf{S}_2|\mathcal{M}_\mathrm{s}^{L},\mathcal{M}^{octree})$ starting from $L=7$ and progressing up to $L=9$. This process aids in addressing the challenges posed by the expansive solution space, gradually refining the optimization for improved convergence.

\noindent{\textbf{\textit{Color Editing}}}. Similar to geometry, vertex colors are also extracted from the volume rendering in a differentiable manner, so users are allowed to manipulate the color of a neural implicit model as well. 
We denote the source color extracted from $\netdensity$ of a pre-trained $\network=\netdensity\circ \netcolor$ as $\mathbf{T}_\mathrm{s}$ and the target color edited by users as $\mathbf{T}_\mathrm{t}$. 
We fix $\netdensity$ and fine-tune $\netcolor$ using a $\mathcal{L}_2$ distance between $\mathbf{T}_\mathrm{s}$ and $\mathbf{T}_\mathrm{t}$ as:
\begin{equation}
    \mathcal{L}_\mathrm{t}=(\mathbf{T}_\mathrm{t}-\mathbf{T}_\mathrm{s})^2.
    \label{eq:color}
\end{equation}

We implement vertex augmentation to enhance editing performance. Usually the edited vertices are a notably small subset of the entire vertex set. In real-world scenarios, artists typically focus on selecting and modifying specific regions of an object to adjust its colors. However, this practice can lead to imbalanced training, where the model encounters very few substantial edits. To tackle this, we introduce an oversampling technique for the edited vertices. This rebalancing method ensures that the edited points constitute at least one-quarter of all the training samples.

We also conduct a camera augmentation. Occlusions may potentially prevent the model from learning visually pleasing results, especially when edited vertices are visible in a sparse set of views. 
Our colored mesh allows for easy augmentation of training views to address this problem. We randomly place $30$ cameras facing edited areas outside the bounding box. This augmentation strategy aims to provide better visibility and coverage of the edited regions, mitigating potential issues caused by occlusions during training.

\section{Experiment}

\begin{figure*}[ht]
\centering
\setlength{\tabcolsep}{0.4\linewidth}
\includegraphics[width=0.9\textwidth]{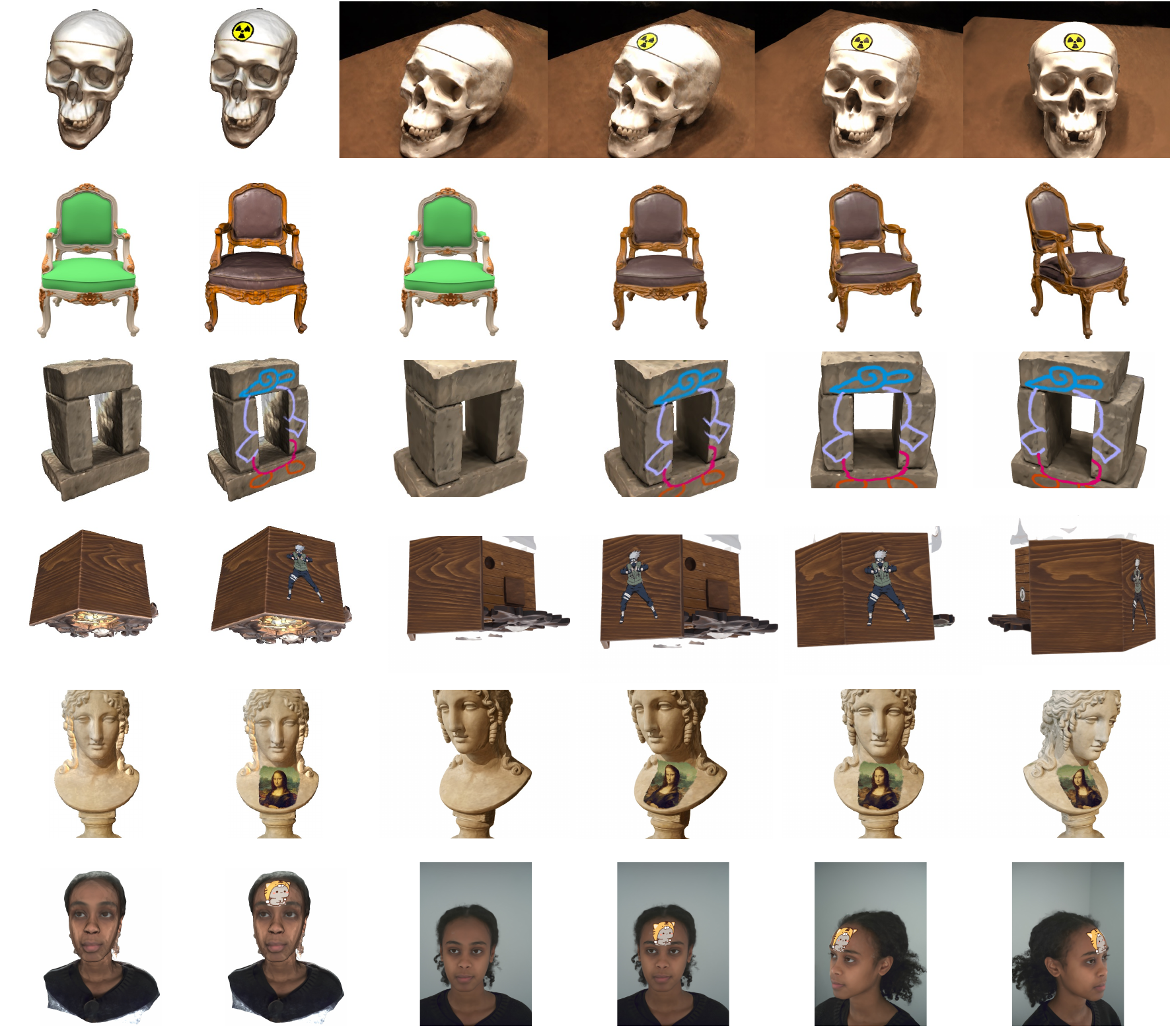}
\begin{small}
\setlength{\tabcolsep}{0\linewidth}
\begin{tabular}{p{0.06\linewidth}
<{\centering}p{0.14\linewidth}
<{\centering}p{0.14\linewidth}<{\centering}p{0.14\linewidth}
<{\centering} p{0.48\linewidth}
<{\centering}p{0.04\linewidth}
<{\centering} }
& Source Mesh & User Edits  & Source  & Rendered Results & 
\end{tabular}
\end{small}
\vspace{-5pt}
\caption{\textbf{Color Editing.}  Color editing results on the colored mesh and their novel views from neural implicit fields. }
\label{fig:te}
\end{figure*}

\begin{figure}[t]
\setlength{\tabcolsep}{0\linewidth}
\includegraphics[width=\linewidth]{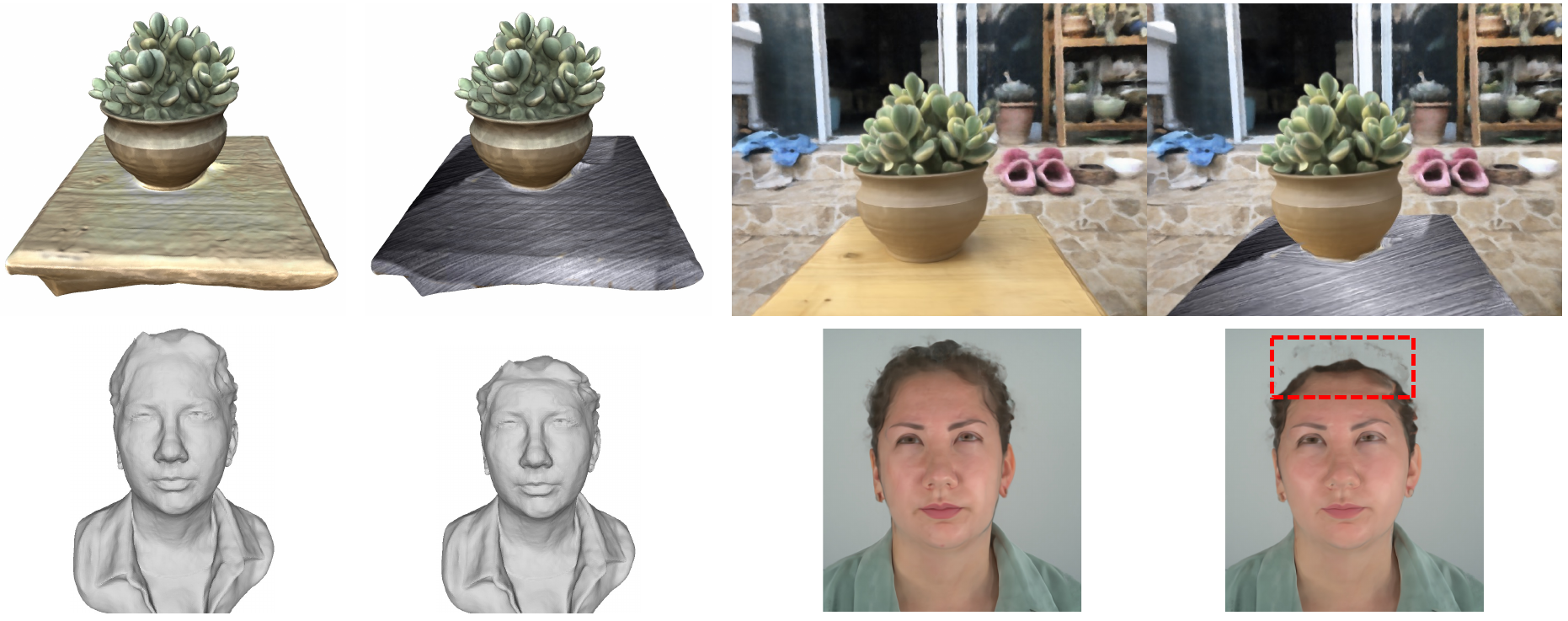}
\begin{small}
\begin{tabular}{p{0.24\linewidth}<{\centering}p{0.24\linewidth}<{\centering}p{0.25\linewidth}<{\centering} p{0.27\linewidth}<{\centering}}
Source Mesh & User Edits & Source & Rendered Results
\end{tabular}
\end{small}
\vspace{-10pt}
\caption{\textbf{Limitations.} Our method does not support direct modifications to a scene's shading and lighting, such as altering the table's shading. Instead, users have to bake these features into the vertex colors. Additionally, it struggles with complex attributes that do not produce a high-quality mesh. For example, deforming non-mesh-reflected elements like hair is not feasible and reports failures (in red).}
\label{fig:limits}
\end{figure}

\subsection{Experiment Setups}

We evaluate our method on three datasets, including the real captured dataset of DTU~\cite{yao2018mvsnet}, the synthetic dataset of NeRF 360°~\cite{mildenhall2020nerf}, and the large scale MVS dataset BMVS~\cite{yao2020blendedmvs}.
For DTU and BMVS, we obtain the pre-trained NeuS~\cite{wang2021neus} models from the official webpage and train new models following the default settings. 
For NeRF 360°, we follow the official split to train a NeuS model with masks. 
All experiments are conducted on these pre-trained models.

For training, we adopt the Adam optimizer with a fixed learning rate of $0.001$.
For geometry and color editing, weights for the Chamfer distance and the color editing losses are $1.0$ and $0.2$, respectively. In geometry editing, we only optimize the density MLP layers. In color editing, we fix the density and only allow the color MLP layers to be optimized.

Our framework supports separate user editings of geometry and color, together with a two-step optimization process, enabling users to sequentially edit them and combine geometry and color alterations. As shown in Fig.~\ref{fig:continue}, users begin by modifying the geometry, adding a star and a magic wand to a smurf, followed by painting on the colored mesh. The differentiable marching tetrahedra are used to optimize the mesh, while our differentiable color extraction function updates the color-related layers within the neural implicit model. Finally, volume rendering generates novel perspectives showcasing edited properties encompassing both geometry and color edits.

\subsection{Ablation Study} 

We begin by evaluating the efficacy of key technical designs in our proposed approach through ablation studies. These studies concentrate on assessing the impact of octree-based editing techniques, the effectiveness of the coarse-to-fine training approach, as well as the contributions made by vertex and camera augmentations.

\noindent{\textbf{\textit{Octree-based Editing Technique}}}. An ablation study is conducted to evaluate the role of the octree in \Fref{fig:abla_octree}. It is observed that a higher resolution of the density grid generally leads to better geometry and color editing outcomes. In \Fref{fig:abla_octree}, a regular grid with $N=128$ showcases more detailed geometric and color characteristics than the regular grid with $N=64$. However, the grid with $N=128$ falls short in generating fine-grained editing results. Attempting to increase the grid size from $N=128$ to $N=256$ results in OOM errors, even on a v100 GPU platform with $80$ GB memory, due to the necessity to calculate and update $256^3=16,777,216$ density nodes via gradient descent optimization. Our octree-based differentiable marching tetrahedra significantly cut down the computational demand during the editing process. With an octree at level $L=9$, the calculation is required for about $1,200,000$ to $2,700,000$ density nodes. This value varies depending on the complexity of the reconstructed scene, as a more intricate scene results in a more complex mesh. We provide the minimum and maximum values observed within our utilized dataset. 
Therefore, when compared to a regular grid of size $N=256$, our method demonstrates better ability to yield more detailed geometric and color attributes using fewer density nodes.

\noindent{\textbf{\textit{Coarse-to-fine Geometry Optimization}}}. In our training methodology, we employ a coarse-to-fine approach, where optimization of the density fields commences at a lower resolution. As the training progresses, we systematically elevate the grid resolution. This strategic training progression, illustrated in \Fref{fig:abla_c2f}, aids significantly in the convergence of the Chamfer loss. Direct training with an excessive number of points can intensify the complexity of this loss, thereby increasing the risk of the model getting caught in local minima. The coarse-to-fine training circumvents this issue, smoothing the path towards more reliable convergence.

\noindent{\textbf{\textit{Vertex and Camera Augmentations}}}. Vertex augmentation helps with imbalanced training with too few vertices being edited, while camera augmentation allows edited vertices to be seen from more viewpoints.
We evaluate both the proposed vertex and camera augmentations in \Fref{fig:abla_va}. Utilizing these augmentations enhances the robustness and accuracy of our color optimization. Without these augmentations, the training process will see fewer substantial vertices with edited colors, resulting in considerably coarser outcomes.

\subsection{Comparisons}

We compare our method with three mesh-guided neural implicit field editing methods, including 
two for general object editing (\textit{i.e.} NeRF-Editing~\cite{yuan2022nerf} and NeuMesh~\cite{yang2022neumesh}), and one for face editing, NeUVF~\cite{ma2022neural}.

In~\Fref{fig:cmp}, we present the comparison between our method and NeRF-Editing as well as NeuMesh. NeRF-Editing and NeuMesh only support mesh-guided deformation with topology fixed~(mesh-guided object motion), while our framework, as depicted in \Fref{fig:result_motion} and \Fref{fig:ge}, accommodates editing for both deformation and topology. Specifically, NeRF-Editing explicitly defines a deformation field between the source and edited meshes to transform the spatial positions of the input template, which does not allow for topology editing. NeuMesh encodes vertices of a prerequisite mesh as shape codes to implicit fields for volume rendering, which only allows for mesh-guided object motion with the topology fixed to the input mesh. In contrast, our method supports mesh-guided object motion and can handle either deformation or topology changes (\Fref{fig:ge}). Therefore, the flexible editing capability of our framework makes it more compatible with existing mesh-based 3D modeling workflows.

NeRF-Editing does not allow for direct color modification, so we only compare with NeuMesh on mesh-guided color painting in \Fref{fig:visualcmp}. Color painting for NeuMesh is performed in a 2D manner, which requires users to paint on a single 2D view and optimizes corresponding color codes under this single-view supervision, which suffers from a major drawback -- it is not intuitive for users to paint on a 2D view compared to editing in 3D perspective, especially when operating on uneven surfaces. 
NeuMesh faces limitations in modeling objects with complex backgrounds due to its reliance on an initial colored mesh, posing challenges in reconstructing intricate backgrounds. Conversely, our method operates via differentiable color extraction in 3D, enabling precise editing of 3D color that seamlessly integrates with radiance fields. In addition, our method can robustly process objects against complex backgrounds, owing to its capability to extract mesh from the neural implicit field inclusive of the background, like the Gundam in \Fref{fig:teaser} and the sculpture in \Fref{fig:ge}.

Moreover, our approach advances both the fine-grained editing results and view consistency. NeuMesh struggles to match the precision of our process in manipulating detailed elements.~\Fref{fig:visualcmp} illustrates that the text and stickers applied by users on NeuMesh yield inconsistent results compared to their intended edits, also exposing inconsistencies in the perspective.
The discrepancy arises from NeuMesh's training method, reliant on a single view, which is more likely to yield inconsistent results. In contrast, our method prioritizes view consistency by implementing edits directly on the mesh. As a result, our approach not only improves the precision of image editing but also enhances the user's visual experience by ensuring a coherent perspective throughout.

NeUVF~\cite{ma2022neural} also falls within the realm of mesh-guided geometry and color editing. However, it is specifically tailored for human faces, as it necessitates the use of facial priors in its training and editing processes. We present visual comparisons between our method and NeUVF~\cite{ma2022neural} on human faces in~\Fref{fig:cmp_neuvf}. Our method not only supports fine-grained edits, such as reshaping a human nose but also enables topology changes, such as removing a girl's hair, which is a feature that NeUVF lacks. NeUVF, built upon NeuTex~\cite{xiang2021neutex}, performs color editing in the UV space, which can result in distorted outcomes, a limitation absent in our method.

\subsection{Results}

In \Fref{fig:ge}, we present additional geometry editing results, showcasing various operations, including topology changes. Our method facilitates various 3D mesh editing operations. Users can manipulate the mesh extracted from a neural implicit model, such as adding new objects, removing vertices and faces, and deforming parts. Our method set the edited mesh as the target, enabling fine-tuning of pre-trained density fields to align with the desired geometry. The results illustrate fine-grained geometry editing, including the addition of a crown to the bear and a bowknot to the dog, removal of the bull's tail and the bird's wing, and non-rigid object deformation. These challenges are often difficult for existing editing methods.

Similar to NeRF-Editing~\cite{yuan2022nerf} and NeuMesh~\cite{yang2022neumesh}, we can also animate an object through skeleton or ARAP (as-rigid-as-possible)~\cite{sorkine2007rigid} mesh deformation while fixing the topology (\Fref{fig:result_motion}). When rendering novel views of a new motion, we define a warping field for each spatial position of the neural implicit fields from the source mesh to the target.

Our method also supports directly editing the color of the original scene, while preserving the geometry. To the best of our knowledge, our work stands as the first method that supports color editing of neural implicit fields driven by a colored mesh, as shown in \Fref{fig:te}. We first extract a colored mesh using our differentiable marching tetrahedra and color extractor.
Users can then edit the color by painting on the mesh using 3D modeling software like Blender and MeshLab. Afterward, the radiance field is finetuned to match the target color. Finally, our method pictures a clear result even when editing on uneven surfaces like the chair with a completely new appearance. Our method offers a user-friendly editing pipeline by directly painting on a colored mesh to control the rendering result. Moreover, our method allows fine-grained color editing, like the delicate painting on the sculpture surface.


\section{Conclusion}

We present a method that edits neural implicit fields using a colored mesh as a guide. This includes developing a differentiable method for extracting colored meshes, enabling the propagation of gradients from the edited mesh to the neural implicit fields. Consequently, when users manipulate the explicit geometry and color, these changes can directly impact the implicit field. Our framework is thus well-suited for interactive editing of neural implicit fields within a mesh-based workflow, compatible with 3D CG software such as Blender and Maya. Furthermore, we introduce an octree-based optimization technique for geometry and color editing. This approach reduces memory usage and facilitates fine-grained editing, enabling detailed adjustments to geometry and color within a 3D space, which is challenging for previous neural implicit field editing methods.

While our approach excels in managing intricate geometric and color details, it does have limitations. Firstly, our method lacks direct support for editing scene shading and lighting. Users must instead bake these features into the vertex colors. This limitation may restrict real-time user control over these aesthetic aspects.
Additionally, our method faces challenges when editing highly intricate structures that cannot produce a high-quality surface mesh, such as human hair. In such scenarios, due to the lack of a reliable underlying structure, our method may struggle to edit the specific part, limiting its applicability in handling intricate geometry.
In the future, we will explore the capability of this differentiable framework in editing the shading and lighting of a scene using techniques such as inverse rendering. We also plan to broaden the applications of our method, including sparse-view neural implicit field reconstruction, which can be achieved by leveraging an initial mesh to differentiably regulate the geometry of neural implicit fields.

\bibliographystyle{ACM-Reference-Format}
\bibliography{ref}

\end{document}

%% file: macros.tex
\usepackage{fp}
\usepackage{expl3}[2012-07-08]
\ExplSyntaxOn
\ExplSyntaxOff

\usepackage{amsmath,amsfonts,bm}

\def\x{{x}}

\def\xi{{\x_i}}

\newcommand{\ignorethis}[1]{}

\def\eqref#1{equation~\ref{#1}}

\def\1{\bm{1}}

\def\vc{{\bm{c}}}
\def\vd{{\bm{d}}}

\def\vn{{\bm{n}}}
\def\vo{{\bm{o}}}
\def\vp{{\bm{p}}}

\def\vr{{\bm{r}}}

\def\vv{{\bm{v}}}

\def\vx{{\bm{x}}}
\def\vy{{\bm{y}}}

\DeclareMathAlphabet{\mathsfit}{\encodingdefault}{\sfdefault}{m}{sl}
\SetMathAlphabet{\mathsfit}{bold}{\encodingdefault}{\sfdefault}{bx}{n}

\DeclareMathOperator{\sign}{sign}

\newcommand{\ignore}[1]{}

\makeatletter
\DeclareRobustCommand\onedot{\futurelet\@let@token\@onedot}
\def\@onedot{\ifx\@let@token.\else.\null\fi\xspace}

\makeatother